\documentclass{aims} % Use the aims.cls file to compile your paper
\usepackage{amsmath}
\usepackage{paralist}
\usepackage{makecell}
\usepackage{mathtools}
\usepackage{cases}
\usepackage{multirow}
\usepackage{subcaption}
\usepackage{amssymb}
\newcommand*{\transpose}{\bgroup\@transpose}
\usepackage[misc]{ifsym}
\usepackage{algorithmic}
\usepackage[linesnumbered,ruled]{algorithm2e}
\newcolumntype{?}{!{\vrule width 1pt}}
\usepackage{epsfig} %For pictures: screened artwork should be set up with an 85 or 100 line screen
\usepackage{epstopdf} %This is to transfer .eps figure to .pdf figure; please compile your article using PDFLaTex or PDFTeXify.
\theoremstyle{definition}

\theoremstyle{remark}
\usepackage[colorlinks=true]{hyperref}
\hypersetup{urlcolor=blue, citecolor=red}
\allowdisplaybreaks

\def\currentvolume{X}
\def\currentissue{X}
\def\currentyear{200X}
\def\currentmonth{XX}
\def\ppages{X-XX}
\def\DOI{10.3934/xx.xxxxxxx}

\theoremstyle{definition}

\title[An overview of differentiable particle filters]
{An overview of differentiable particle filters for data-adaptive sequential Bayesian inference} % Only the first word and proper nouns should be capitalized.

\author[Xiongjie Chen and Yunpeng Li]{}


\thanks{$^*$Corresponding author: Xiongjie Chen}
\begin{document}
	\maketitle
	
	% Enter the first author's name and email address; email addresses are required for each author.
	% Use footnote notations to indicate address and affiliations with commas between numbers if more than one address applies; see below for examples.
	\centerline{\scshape
		Xiongjie Chen$^{{\href{xiongjie.chen@surrey.ac.uk}{\textrm{\Letter}}}*1}$
		and Yunpeng Li$^{{\href{yunpeng.li@surrey.ac.uk}{\textrm{\Letter}}}1}$}
	
	\medskip
	
	{\footnotesize
		% Enter the full affiliation and country name:
		% Do not insert commas or periods at the end of lines.
		\centerline{$^1$University of Surrey, United Kingdom}
	} % Do not forget to end {\footnotesize with the sign }
	\bigskip
	
	%%%%%%%%%%%%%%%%%%%%%%%%%%%%%%%%%%%%%%%%%%%%%%%%%%%%%%%
	%             5. ABSTRACT
	%%%%%%%%%%%%%%%%%%%%%%%%%%%%%%%%%%%%%%%%%%%%%%%%%%%%%%%
	
	\begin{abstract}
		By approximating posterior distributions with weighted samples, particle filters (PFs) provide an efficient mechanism for solving non-linear sequential state estimation problems. While the effectiveness of particle filters has been recognised in various applications, their performance relies on the knowledge of dynamic models and measurement models, as well as the construction of effective proposal distributions. An emerging trend involves constructing components of particle filters using neural networks and optimising them by gradient descent, and such data-adaptive particle filtering approaches are often called differentiable particle filters. Due to the expressiveness of neural networks, differentiable particle filters are a promising computational tool for performing inference on sequential data in complex, high-dimensional tasks, such as vision-based robot localisation. In this paper, we review recent advances in differentiable particle filters and their applications. We place special emphasis on different design choices for key components of differentiable particle filters, including dynamic models, measurement models, proposal distributions, optimisation objectives, and differentiable resampling techniques.
	\end{abstract}
	
	%%%%%%%%%%%%%%%%%%%%%%%%%%%%%%%%%%%%%%%%%%%%%%%%%%%%%%
	%                   6. BODY
	%%%%%%%%%%%%%%%%%%%%%%%%%%%%%%%%%%%%%%%%%%%%%%%%%%%%%%
	
	% Only the first word and proper nouns of section titles should be capitalized.
	% The title of section 1:
	\section{Introduction}
	\label{sec:introduction}
	In many signal processing and control problems, we are often interested in inferring posterior distributions of latent variables. These variables cannot be directly observed but manifest themselves through an observation process conditioned on the unobserved variables. The posterior distribution of latent variables is defined by two system elements, i.e. a prior distribution of the unobserved variables and a likelihood function that describes the relationship between latent variables and observations. When dealing with sequential data for both latent variables and observations, this task is known as the Bayesian filtering problem, where one recursively updates the posterior distribution of latent variables. In simple systems, such as linear-Gaussian state-space models, the evolving posterior distribution can be exactly derived with analytical expressions using Kalman filters~\cite{kalman1960new}. However, real-world problems often involve non-linear and non-Gaussian systems, where posterior distributions are not analytically available due to the computation of intractable high-dimensional integrals. As a result, many approximation-based approaches have been proposed, including the extended Kalman filter~\cite{anderson1979optimal,daum2015extended}, the unscented Kalman filter~\cite{wan2000unscented}, and grid-based filters~\cite{bucy1971digital}. Despite these developments, they often perform poorly in high-dimensional, highly non-linear cases~\cite{doucet2001introduction}.
	
	One popular alternative solution to the non-linear and non-Gaussian filtering problem is particle filters, also known as sequential Monte Carlo (SMC) methods~\cite{gordon1993novel, djuric2003particle,doucet2001sequential}. In particle filters, the target posterior distributions are recursively approximated by a sequence of empirical distributions associated with a set of particles, i.e. weighted samples. Particle filters do not assume linearity or Gaussianity on the considered state-space model, and convergence results on particle filtering methods have been established~\cite{del2000branching,crisan2002survey,del1998measure,elvira2017adapting,elvira2021performance}. Due to these favourable properties, a variety of particle filtering methods have been developed following the seminal work on the bootstrap particle filter (BPF) introduced in~\cite{gordon1993novel}. Examples include auxiliary particle filters (APFs)~\cite{pitt1999filtering,elvira2019elucidating,branchini2021optimized} to guide particles into regions where the posterior distribution has higher densities. The Gaussian sum particle filters (GSPFs)~\cite{kotecha2003gaussian,kotecha2001gaussian,kotecha2003gaussian_particle} approximate posterior distributions with Gaussian mixture models by building banks of Gaussian particle filters. The unscented particle filters (UPFs)~\cite{van2000unscented,rui2001better,julier2004unscented} incorporate the latest observations in the construction of proposal distributions generated by unscented Kalman filters (UKFs). Several other variants of particle filters, including Rao-Blackwellised particle filters~\cite{doucet2000rao,de2002rao}, multiple particle filters~\cite{djuric2007, djuric2013}, particle flow particle filters~\cite{bunch2016approximations, li2017particle,heng2021gibbs}, have been designed for more efficient sampling in high-dimensional spaces. Although these variants of particle filters have shown to be effective in many applications~\cite{zhang2017multi,malik2011particle,gunatilake2022novel,van2019particle,pozna2022hybrid}, practitioners often have to design hand-crafted transition kernels of latent states and likelihood functions, which is often time-consuming and relies on extensive domain knowledge. 
	
	To address this challenge, various techniques have been proposed for parameter estimation in state-space models and particle filters~\cite{kantas2009overview,kantas2015particle}. The particle Markov chain Monte Carlo (PMCMC) method and its variants provide a mechanism to perform
	parameter estimation for sequential Monte Carlo methods using Markov chain Monte Carlo (MCMC) sampling~\cite{andrieu2010particle,lindsten2014particle}. The $\text{SMC}^2$ algorithm~\cite{chopin2013smc2} is a particle equivalent of the PMCMC method, which simultaneously tracks the posterior of model parameters and latent variables using two layers of particle filters. Based on a similar nested filtering framework as in the $\text{SMC}^2$ algorithm, a series of nested hybrid filters were proposed within a purely recursive structure and thus are more suitable for jointly inferring both static parameters of the system and latent posteriors in on-line settings~\cite{perez2018probabilistic,crisan2018nested,perez2021nested}. In~\cite{chouzenoux2020graphem,elvira2022graphical,chouzenoux2023sparse}, the transition matrix and the covariance matrix in linear-Gaussian state-space models were interpreted as a directed graph, and a GraphEM method based on the expectation-maximisation (EM) algorithm was proposed to learn the directed graph.
	
	An emerging trend in this active research area is to construct and learn components of particle filters using machine learning techniques, such as neural networks and gradient descent. Such particle filters are often named differentiable particle filters (DPFs)~\cite{jonschkowski18,karkus2018particle,wen2021end,chen2021differentiable,chen2022conditional,corenflos2021differentiable}. In particular, differentiable particle filters construct their dynamic models, measurement models, and proposal distributions using expressive neural networks that can model complex high-dimensional non-linear non-Gaussian systems. One main obstacle to the development of fully-differentiable particle filters is the resampling step which is known to be non-differentiable in classical resampling methods. To this end, different resampling strategies have been proposed to address this issue, e.g. no resampling~\cite{jonschkowski18}, soft resampling~\cite{karkus2018particle}, entropy-regularised optimal transport resampling~\cite{corenflos2021differentiable}, and transformer-based resampling~\cite{zhu2020towards}. Another key ingredient to developing data-adaptive particle filters is a loss function, which is minimised using gradient descent to optimise parameters of neural networks used in the construction of differentiable particle filters. 
	
	Despite the surging interest in differentiable particle filters, as far as we know, there is an absence of a comprehensive review of recent advances in this field. Therefore, the main objective of this paper is to provide an overview of the design of differentiable particle filters. We place special emphasis on different design choices of key ingredients in differentiable particle filters, and aim to answer the following questions:
	\begin{itemize}
		\item How to model the transition of latent states in state-space models with neural networks?
		\item How to estimate the likelihood of observations given states using neural networks?
		\item How can we construct proposal distributions that lead to better approximations of the true posterior in complex environments?
		\item What loss functions should we use when training differentiable particle filters?
		\item How to perform resampling for differentiable particle filters?
	\end{itemize}
	
	The remainder of this paper is organised as follows. In Section~\ref{sec:preliminaries}, we provide the necessary background knowledge for introducing differentiable particle filters. In Section~\ref{sec:sampling_distribution}, we detail different design choices of sampling distributions of latent states in differentiable particle filters, i.e. their dynamic models and proposal distributions. Section~\ref{sec:measurement_model} discusses the construction of differentiable particle filters' measurement models. In Section~\ref{sec:resampling}, we list examples of resampling schemes used in differentiable particle filters. Loss functions used for optimising differentiable particle filters are introduced in Section~\ref{sec:loss_function}. We provide implementation details of some differentiable particle filters in Section~\ref{sec:examples}. We conclude the paper in Section~\ref{sec:conclusion}.
	
	\section{Preliminaries}
	\label{sec:preliminaries}
	\subsection{State-Space Models}
	In this section, we detail the problem we consider in this paper, including a brief introduction to state-space models and the general goal of Bayesian filtering approaches.
	
	State-space models (SSMs) refer to a class of sequential models that consist of two discrete-time random sequences, the latent state sequence $({x}_t)_{t\geq0}$ defined on $\mathcal{X}\subseteq\mathbb{R}^{d_\mathcal{X}}$, and the observed measurement sequence $({y}_t)_{t\geq1}$ defined on $\mathcal{Y}\subseteq\mathbb{R}^{d_\mathcal{Y}}$. The latent state sequence $({x}_t)_{t\geq0}$  is characterised by a Markov process with an initial distribution $\pi(x_0)$ and a transition kernel $p(x_{t} | {x}_{t-1})$ for $t\geq 1$. The observation ${y}_t$ is conditionally independent given the current latent state $x_t$. In this review, we focus on parameterised state-space models defined as follows:
	\begin{align}
            \label{eq:ssm_1}
		&{x}_{0} \sim \pi(x_0; \theta)\,, \\
		\label{eq:ssm_2}
            &{x}_{t}| x_{t-1} \sim p(x_{t}| {x}_{t-1}; \theta) \text { for } t \geq 1 \,,\\
            \label{eq:ssm_3}
		&{y}_{t}| x_t \sim p(y_t | {x}_{t} ; \theta) \text { for } t \geq 1\,,
	\end{align}
	where $\theta\in\mathbb{R}^{d_\theta}$ is the parameter set of interest, $\pi(x_0\, ; \theta)$ is the initial distribution of the latent state, $p(x_{t}| {x}_{t-1}; \theta)$ is the dynamic model that describes the transition of the latent state, and $p(y_t | {x}_{t};\theta)$ is the measurement model estimating the conditional likelihood of observation $y_t$ given $x_t$. Denote by ${x}_{0:t}:=\{{x}_0,\,\,\cdots,\,\,{x}_t\}$ and ${y}_{1:t}:=\{{y}_1,\,\,\cdots,\,\,{y}_t\}$ respectively the sequence of latent states and the sequence of observations up to time step $t$, the parameterised state-space model defined by Eqs.~\eqref{eq:ssm_1}~-~\eqref{eq:ssm_3} can also be defined through the following joint distribution of $x_{0:t}$ and $y_{1:t}$:
	\begin{equation}
		p\left(x_{0: t}, y_{1: t};\theta\right)=\pi\left(x_0;\theta\right)\prod_{k=1}^{t} p\left(x_{k} | x_{k-1};\theta\right) p\left(y_{k} | x_{k};\theta\right)\,,
	\end{equation}
	with $p\left(x_{0}, y_{1: 0};\theta\right):=\pi\left(x_{0};\theta\right)$.
 
    In Bayesian filtering problems, the goal is to infer recursively in time the joint posterior distribution $p\left(x_{0: t}| y_{1: t};\theta\right)$, the marginal posterior distribution $p\left(x_{t}| y_{1: t};\theta\right)$, or the expectation $\mathbb{E}_{p\left(x_{0: t}| y_{1: t};\theta\right)}\big[\psi_t{(x_{0:t})}\big]$ of a function $\psi_t{(\cdot)}:\mathcal{X}^{t+1}\rightarrow\mathbb{R}^{d_{\psi_t}}$ w.r.t. the joint posterior distribution or the log evidence $L_t(\theta):=\log p(y_{1:t};\theta)$.
    
    We use the joint posterior and the log evidence as two examples to show how they can be updated recursively, other quantities of interest can be derived similarly. At the $t$-th time step, the joint posterior distribution $p\left(x_{0: t}| y_{1: t};\theta\right)$ can be written as follows:
	\begin{align}
		\label{eq:joint_posterior_original}
		p\left(x_{0: t}| y_{1: t};\theta\right)&=\frac{p\left(y_{1: t}|x_{0: t};\theta\right)p\left(x_{0: t};\theta\right)}{p(y_{1:t};\theta)}\\
		&=\frac{p\left(y_{1: t}|x_{0: t};\theta\right)p\left(x_{0: t};\theta\right)}{\int_{\mathcal{X}^{t+1}}\,p(x_{0:t},y_{1:t};\theta)\text{d}x_{0:t}}\,,
	\end{align}
    with $p(x_0|y_{1:0};\theta):={\pi(x_0;\theta)}$. In addition to Eq.~\eqref{eq:joint_posterior_original}, which directly computes the joint posterior by applying Bayes' theorem, we can also obtain a recursive formula for the joint posterior $p\left(x_{0: t}| y_{1: t};\theta\right)$:
	\begin{align}
        \label{eq:recursive_update_posterior_1}
		&p(x_0|y_{1:0};\theta):={\pi(x_0;\theta)}\,,\\
		\nonumber\\
		&p(x_{0:t}| y_{1:t-1};\theta)= p(x_{0:t-1}| y_{1:t-1};\theta)p(x_t| x_{t-1};\theta)\,,\;t\geq 1\,,\\
		\nonumber\\
		&p(x_{0:t}| y_{1:t};\theta)
		=\frac{p(y_t| x_t;\theta)p(x_{0:t}| y_{1:t-1};\theta)}{p(y_t| y_{1:t-1};\theta)}\\
		&\phantom{p(x_{0:t}| y_{1:t};\theta)}=p(x_{0:t-1}| y_{1:t-1};\theta)\frac{p(y_t| x_t;\theta)p(x_t| x_{t-1};\theta)}{p(y_t| y_{1:t-1};\theta)}\,,\; t\geq 1\,,
        \label{eq:recursive_update_posterior_2}
	\end{align}
        with $p(y_1| y_{1:0};\theta):=p(y_1;\theta)$. 
	The log evidence $L_t(\theta):=\log p(y_{1:t};\theta)$ also satisfies the following recursion:
	\begin{gather}
		\label{eq:obs_marg_lki}
            L_t(\theta)=\sum_{k=1}^{t} l_t(\theta)=L_{t-1}(\theta)+l_t(\theta)\,,\\
		l_t(\theta):=\log p(y_t|y_{1:t-1};\theta)=\log\int_{\mathcal{X}^{t+1}} p(y_t| x_{t}; \theta)p(x_{0:t}| y_{1:t-1};\theta)\text{d}x_{0:t}\,.
        \label{eq:obs_marg_lki_2}
	\end{gather} 
    However, except for only a limited class of state-space models such as linear-Gaussian models, obtaining $p(x_{0:t}| y_{1:t};\theta)$ and $\log p(y_{1:t};\theta)$ from $p(x_{0:t-1}| y_{1:t-1};\theta)$ and $\log p(y_{1:t-1};\theta)$ using Eqs.~\eqref{eq:recursive_update_posterior_1} --~\eqref{eq:obs_marg_lki_2} are usually analytically intractable since they involve the computation of complex high-dimensional integrals over $\mathcal{X}^{t+1}$. As an alternative, particle filters resort to Monte Carlo integration techniques to approximate the posterior distributions and other quantities of interest. We present more details about particle filters in the next section.
	\subsection{Particle Filters}
	Particle filters (PFs), also known as sequential Monte Carlo (SMC) methods, are a family of Monte Carlo algorithms designed for approximating the latent posterior distributions sequentially in state-space models~\cite{doucet2001introduction}. In particular, particle filters approximate the analytically intractable joint posterior $p(x_{0:t}| y_{1:t};\theta)$ with an empirical distribution consisting of a set of weighted samples $\{W_t^i,x_{0:t}^i\}_{i\in [N_p]}$:
	\begin{equation}
		\label{eq:pf_approx}
		p(x_{0:t}| y_{1:t};\theta)\approx\sum_{i=1}^{N_p}W_t^i\, \delta_{x_{0:t}^i}(x_{0:t})\,,
	\end{equation}
	where $[N_p]:=\{1,\cdots,N_p\}$, $N_p$ is the number of particles, $\delta_{x_{0:t}^i}(\cdot)$ denotes the Dirac delta function located in $x_{0:t}^i$, and $W_t^i\geq 0$ with $\sum_{i=1}^{N_p}W_t^i=1$ is the normalised importance weight of the $i$-th particle at the $t$-th time step. 
	%	The weight $W_t^i$ indicates the importance of the $i$-th state sequence $x_{0:t}^i$, i.e., particles with higher importance weights are believed to be closer to the true state than those with lower importance weights. 
	By representing the intractable posterior distributions of latent variables with sets of weighted samples, particle filters can provide efficient and asymptotically unbiased estimates of the posterior distributions, i.e. they are able to approximate the posterior distributions arbitrarily well under weak assumptions when the number of particles $N_p\rightarrow +\infty$~\cite{del2000branching,crisan2002survey,del1998measure,elvira2017adapting,elvira2021performance}.
	
	As an importance sampling-based algorithm~\cite{elvira2021advances,hesterberg1995weighted,tokdar2010importance,agapiou2017importance}, the performance of particle filters highly relies on the distribution where particles are sampled from, which is often called the proposal distribution. An ideal proposal distribution should allow efficient sampling of particles and induce tractable importance weights. In bootstrap particle filters (BPFs), the proposal distribution is simply the system dynamic $p(x_{t}| x_{t-1};\theta)$~\cite{gordon1993novel}. Denote by $w_t^i$ unnormalised importance weights of particles, in bootstrap particle filters, particle weights are updated according to the likelihood of observation $y_t$ given state $x_t^i$:
	\begin{equation}
		w_t^i= w_{t-1}^i p(y_t| x_t^i;\theta)\,,
	\end{equation}
	with $w_{0}^i=1$. An obvious drawback of bootstrap particle filters is that sampling from the system dynamic $p(x_{t}| x_{t-1};\theta)$ often results in trivially small weights for the majority of particles, especially when dealing with long sequences. This is known as the weight degeneracy problem, where the Monte Carlo approximation of the posterior is dominated by only a few particles with large weights~\cite{bickel2008sharp}. Weight degeneracy leads to poor estimates of posteriors, and most of the computational cost will be wasted on particles with negligible weights. Therefore, to obtain better approximations of the posterior, information from the latest observations is usually utilised in the construction of proposal distributions, which gives rise to the guided particle filters~\cite{chopin2020introduction}. Compared to bootstrap particle filters, by selecting an appropriate proposal distribution, guided particle filters can propose particles located in regions where the posterior has higher densities and thus yield more uniformly distributed particle weights. Denote the proposal distributions by $q(x_{t}| y_t, x_{t-1};\phi)$ for $t\geq1$, in guided particle filters, the unnormalised importance weight of the $i$-th particle is updated recursively through:
	\begin{equation}
		w_t^i= w_{t-1}^i \frac{p(y_t| x_t^i;\theta)p(x_t^i| x_{t-1}^i;\theta)}{q(x_{t}^i| y_t, x_{t-1}^i;\phi)}\,,
	\end{equation}
	with $w_{0}^i=1$. 
	
	However, even with proposal distributions constructed with information from observations, the weight degeneracy problem is still a challenging issue since the dimension of the joint posterior $p(x_{0:t}| y_{1:t};\theta)$ increases over time. Therefore, particle resampling~\cite{li2015resampling_b}, a critical step of particle filters, was introduced to further mitigate the weight degeneracy problem. It has been proved in the literature, both empirically and theoretically, that resampling steps play a crucial role in avoiding the weight degeneracy problem and stabilising the Monte Carlo error over time~\cite{gerber2019negative, li2015resampling}.
 
 There are different choices of resampling schemes, including multinomial resampling, residual resampling, stratified resampling, and systematic resampling~\cite{douc2005comparison,bolic2004resampling}. Resampling is often triggered when a pre-defined condition is met. One commonly used threshold metric for resampling is the effective sample size (ESS). The effective sample size in particle filters can be interpreted as the number of particles needed from the true posterior to produce the same variance in its estimates as that of the $N_p$ samples drawn from the proposal distribution~\cite{doucet2000sequential}. Since the actual value of the ESS is intractable in general, an approximation of the ESS is often used instead:
	\begin{equation}
        \label{eq:ESS}
		\operatorname{ESS}_t\left(\left\{W_{t}^{i}\right\}_{i \in\left[N_{p}\right]}\right)=\frac{1}{\sum_{i=1}^{N_{p}}\left(W_{t}^{i}\right)^{2}}\,.
	\end{equation}
     When using the effective sample size as the threshold metric, resampling steps are triggered only when $\operatorname{ESS}_t\leq \text{ESS}_{\text{min}}$, where $\text{ESS}_{\text{min}}$ refers to a predefined threshold for effective sample sizes. While the performance of particle filters is not sensitive to the value of $\text{ESS}_{\text{min}}$,  one common choice is to set $\text{ESS}_{\text{min}}=\frac{N_p}{2}$. Notably, it has been shown that the estimator of the ESS defined by Eq.~(16) provides an accurate estimation of the true ESS only in specific cases and may fail to detect certain issues with the particle approximation~\cite{elvira2022rethinking,martino2017effective}. As a result, alternatives to Eq.~(16) have recently been proposed~\cite{elvira2022rethinking,martino2017effective,huggins2015convergence,li2020flexible}.
     
     A pseudocode for particle filtering approaches is provided in Algorithm~\ref{SMC}. Note that the presented pseudocode is a standard implementation of particle filters. Some variants of particle filtering approaches may not fit in the framework, e.g. auxiliary particle filters~\cite{pitt1999filtering,elvira2019elucidating,branchini2021optimized} and Rao-Blackwellised particle filters~\cite{doucet2000rao,de2002rao}. However, we do not extend it to more generic particle filtering algorithms because most existing differentiable particle filtering frameworks follow the structure of Algorithm~\ref{SMC}.
	
	\begin{algorithm}[t]
		\begin{algorithmic}[1]
			\caption{Pseudocode for standard particle filtering algorithms.}\label{SMC}
			\STATE \textbf{Require}: A state-space model, resampling threshold $\text{ESS}_{\text{min}}$, particle number $N_p$, proposal distributions $q(x_{t}| y_t, x_{t-1};\phi)$, a resampling scheme $\mathcal{R}(\{W_t^i\}_{i\in[N_p]})$ that outputs indices $A_t^i$ of selected particles;
			\STATE \textbf{Initialisation}: Sample $x_0^i\overset{\text{i.i.d}}{\sim}\pi(x_0;\theta)$ for $\forall i\in[N_p]$;
			\STATE Set weights $w_0^i=1$ for $\forall i\in[N_p]$;
			\STATE Set normalise weights $W_0^i=\frac{1}{N_p}$ for $\forall i\in[N_p]$;
			\STATE Set $L_0(\theta)=1$;
			\STATE Set the effective sample size $\text{ESS}_0(\{W_{0}^i\}_{i\in [N_p]})=\frac{1}{\sum_{i=1}^{N_p}(W_{0}^i)^2}=N_p$;
			\FOR {$t=1$ to $T$}
			
			\IF {$\text{ESS}_t(\{W_{t-1}^i\}_{i\in [N_p]}) < \text{ESS}_{\text{min}}$} 
			\STATE $A_t^{1:N_p}\leftarrow$ $\mathcal{R}(\{W_{t-1}^i\}_{i\in[N_p]})$;
			\STATE $\tilde{{w}}_{t-1}^i\leftarrow {1}$ for $\forall i \in[N_p]$;
			\ELSE
			\STATE $A_t^{i}\leftarrow i$ for $\forall i \in[N_p]$;
			\STATE $\tilde{{w}}_{t-1}^i\leftarrow {{w}}_{t-1}^i$ for $\forall i \in[N_p]$;
			\ENDIF
			\STATE Sample $x_t^i\overset{\text{i.i.d}}{\sim}q(x_t| y_t, x^{A_t^i}_{t-1};\phi)$ for $\forall i\in[N_p]$;
			\STATE $w_t^i=\tilde{w}_{t-1}^i \frac{p(y_{t}| x_{t}^{i}; \theta) p(x_t^i| x_{t-1}^{A_t^i};\theta)}{q(x_t^i| y_{t},x_{t-1}^{A_t^i};\phi)}$;
			\STATE (optional) Estimate the log-likelihood of observations $L_t(\theta)$:\\ $l_t(\theta)=\log\frac{\sum_{i=1}^{N_p}w_t^i}{\sum_{i=1}^{N_p}\tilde{w}_{t-1}^i}$, $L_t(\theta)=L_{t-1}(\theta)+l_t(\theta)$;
			\STATE Normalise weights $W_t^i={w_t^i}/{\sum_{n=1}^{N_p}w_t^n}$ for $\forall i\in[N_p]$;
			\STATE Compute the effective sample size: $\text{ESS}_t(\{W_{t}^i\}_{i\in [N_p]})=\frac{1}{\sum_{i=1}^{N_p}(W_{t}^i)^2}$;
			% 		\ELSE
			% 		\STATE  $s_t^i=\tilde{s}_t^i$ for $i=1,...,N_p$;
			\ENDFOR			
		\end{algorithmic}
	\end{algorithm}
	\subsection{Differentiable Particle Filters}
	\label{subsec:intro_dpfs}
	While parameter estimation techniques for particle filters have long been an active research area~\cite{malik2011particle, doucet2003parameter, andrieu2005line}, many existing approaches are restricted to scenarios where a subset of model parameters is known or the model has to follow certain structures~\cite{kantas2015particle}. Recently, an emerging trend in this direction is the so-called differentiable particle filters (DPFs)~\cite{jonschkowski18,karkus2018particle,wen2021end,corenflos2021differentiable,zhu2020towards,kloss2021train,rosato2022efficient}. The main idea of differentiable particle filters is to develop data-adaptive particle filters by constructing particle filters' components through neural networks, and optimise them by minimising a loss function through gradient descent. With the expressiveness of neural networks, differentiable particle filters have shown their effectiveness in complicated applications, e.g. vision-based robot localisation, where the observation can be high-dimensional unstructured data like images~\cite{jonschkowski18,karkus2018particle,wen2021end,corenflos2021differentiable}. We provide in the remainder of this section an introduction of key components in differentiable particle filters, before we delve into more detailed discussions in the following sections. 
	
	In differentiable particle filters, given a particle $x_{t-1}^i$ at the $(t-1)$-th time step, we can generate a new particle $x_t^i$ for the $t$-th time step as follows~\cite{jonschkowski18}:
	\begin{equation}
		\label{eq:dpf_dynamic_model}
		x_t^i=g_\theta(x_{t-1}^i, \alpha_t^i)\sim p(x_t| x_{t-1}^i;\theta)\,,
	\end{equation}
	where $\alpha_t^i\in\mathbb{R}^{d_\alpha}$ simulates process noise, $g_\theta(\cdot): \mathcal{X}\times \mathbb{R}^{d_\alpha}\rightarrow \mathcal{X}$ is a function that is differentiable w.r.t. both $x_{t-1}^i$ and $\alpha_t^i$. Regarding the measurement model, several differentiable particle filters estimate the conditional likelihood of an observation $y_t$ given $x_t^i$ as follows~\cite{jonschkowski18,karkus2018particle,wen2021end,corenflos2021differentiable}:
	\begin{equation}
		\label{eq:dpf_measurement_model}
		p(y_t| x_t^i;\theta)\propto l_\theta(y_t, x_t^i)\,,
	\end{equation}
	where $l_\theta(\cdot):\mathcal{Y}\times\mathcal{X}\rightarrow\mathbb{R}$ is a function that is differentiable w.r.t. both $y_t$ and $x_t^i$. In addition, proposal distributions can be specified through a function $f_\phi(\cdot):\mathcal{X}\times \mathcal{Y}\times\mathbb{R}^{d_\beta}\rightarrow\mathcal{X}$ that are differentiable w.r.t. $x_{t-1}^i$, $y_t$, and $\beta_t^i$~\cite{chen2021differentiable}:
	\begin{equation}
		x_t^i=f_\phi(x_{t-1}^i,y_t, \beta_t^i)\sim q(x_{t}^i| y_t, x_{t-1}^i;\phi)\,,
	\end{equation}
	where $\beta_t^i\in\mathbb{R}^{d_\beta}$ is the noise term used to generate proposal particles.

    As a neural network-based machine learning model, the components of differentiable particle filters need to be trained often by minimising a given loss function $\mathcal{L}(\theta, \phi)$, e.g. the prediction error between the predicted latent states and the ground-truth latent states if they are available or the evidence lower bound (ELBO) of observation log-likelihoods. To minimise $\mathcal{L}(\theta, \phi)$ via gradient descent, one needs to compute the gradient of $\mathcal{L}(\theta, \phi)$ w.r.t. the system parameter $\theta$ and the proposal parameter $\phi$, and we call a component differentiable if it can be interpreted as a differentiable function, i.e. it has a deterministic output and the derivative of its output w.r.t. its input exists at each point in its domain -- it is allowed in a machine learning context to have finitely many points where the derivatives do not exist. However, neither generating new particles nor the resampling steps in particle filters are differentiable in their vanilla forms, as they both lead to stochastic outputs. In addition, the output of the resampling step changes in a discrete manner as small changes in its input (particle weights) can lead to discrete changes in its output (particle indices). We introduce below how to construct differentiable components for particle filters such that the parameters $\theta$ and $\phi$ can be optimised using gradient descent.
    
    Although they are not specifically designed for differentiable particle filters, different strategies have been proposed to achieve differentiable sampling, including the REINFORCE gradient estimator~\cite{williams1992simple} and the reparameterisation trick~\cite{kingma2014auto}, among which the reparameterisation trick is the one that most commonly used in differentiable particle filters. The essence of the reparameterisation trick is to rewrite samples from the target distribution as the output of a deterministic and differentiable function parameterised by distribution parameters such that the gradient of samples w.r.t. the parameters exists. In particular, consider a target distribution $\mu_\beta$ parameterised by $\beta$ and defined on $\mathbb{R}^{d_x}$, the reparameterisation trick rewrites  $\mu_\beta$ as a push-forward probability measure $\mu_\beta=G_\beta\#\nu$ of a base distribution $\nu$ defined on $\mathbb{R}^{d_\epsilon}$, where $G_\beta(\cdot):\mathbb{R}^{d_\epsilon}\rightarrow\mathbb{R}^{d_x}$ is a differentiable deterministic function parameterised by $\beta$. Samples following the target distribution $\mu_\beta$ can be obtained via sampling from the base distribution $\nu$ and transforming base samples $\epsilon\sim\nu$ through $x=G_\beta(\epsilon)$, with $x\sim\mu_\beta$. The gradient of an expectation of a function $\psi(\cdot):\mathbb{R}^{d_x}\rightarrow \mathbb{R}^{d_\psi}$ can now be rewritten as:
    \begin{align}
        \nabla_\beta\mathbb{E}_{\mu_\beta}[\psi(x)]&=\nabla_\beta\mathbb{E}_{\nu}[\psi(G_\beta(\epsilon))]\\
        &=\mathbb{E}_{\nu}[\nabla_\beta \psi(G_\beta(\epsilon))]\\
        &\approx \frac{1}{K}\sum_{k=1}^{K}\nabla_\beta \psi(G_\beta(\epsilon_k))\,,
    \end{align}
    where $\epsilon_k \overset{i.i.d}{\sim} \nu$ are i.i.d samples from $\nu$, and $K$ is the number of samples used to estimate the required gradient. Gradient estimators obtained by using the reparameterisation trick were shown to be unbiased~\cite{kingma2014auto}. Note that not all distributions are reparameterisable. Some examples of reparameterisable distributions, e.g. Gaussian distributions, are listed in~\cite{kingma2014auto}. 
	
	%	A more general way to compute gradients of expectations of a function w.r.t. distribution parameters is the REINFORCE gradient estimator~\cite{williams1992simple}, which has demonstrated its effectiveness in reinforcement learning applications~\cite{sutton1999policy,peters2006policy,agarwal2021theory}. , which typically has less variance than the REINFORCE estimator~\cite{kingma2014auto}.

	As for the resampling step, it is inherently non-differentiable because of the stochastic and discrete nature of multinomial resampling. Specifically, when using the inverse cumulative distribution function (CDF) algorithm to resample particles according to their importance weights, a small change in weights can lead to discrete changes in the resampling output. Recent efforts in differentiable particle filters have proposed various solutions to address this issue~\cite{jonschkowski18,karkus2018particle,corenflos2021differentiable,zhu2020towards,rosato2022efficient}. Several differentiable particle filtering approaches bypass resampling steps in gradient computation or simply do not use resampling~\cite{jonschkowski18,karkus2018particle}. It was proposed in~\cite{jonschkowski18} to avoid backpropagating gradients across time steps, and the whole sequence was decomposed into multiple independent steps by truncating backpropagation after every single time step. It was suggested in~\cite{karkus2018particle} to use the same particles as the resampling output, i.e. no resampling, and this strategy was found to be effective in low uncertainty environments. In standard resampling steps, the weights of survived particles $\tilde{w}_t^i$ are set to be a constant, implying the gradient $\frac{\partial\tilde{w}_t^i}{\partial {w}_t^i}$ is always zero. To prevent this, several resampling schemes are proposed for differentiable particle filters to generate non-zero gradients $\frac{\partial\tilde{w}_t^i}{\partial {w}_t^i}$. For instance, in~\cite{rosato2022efficient}, the sum of unnormalised weights after resampling was kept the same as before resampling:
	$\tilde{w}_t^i=\frac{1}{N_p}\sum_{j=1}^{N_p}w_t^j$,
	such that the gradient $\frac{\partial\tilde{w}_t^i}{\partial {w}_t^j}$ becomes non-zero. 
    % However, this resampling technique still yields zero gradients when using loss functions that only involve normalised weights. 
    In~\cite{karkus2018particle}, it was proposed to resample particles from a multinomial distribution designed as a linear interpolation between the original multinomial distribution and a multinomial distribution with uniform weights. Biases caused by resampling from this interpolation are corrected by modifying the importance weights after resampling as a function of the original weights. This implies that the importance weights of particles after resampling now depend on the importance weights before resampling, hence $\frac{\partial\tilde{w}_t^i}{\partial {w}_t^i}$ becomes non-zero.
	
	However, the aforementioned resampling techniques do not lead to fully differentiable resampling because they still rely on multinomial resampling, which is inherently non-differentiable as explained above. One solution to fully differentiable particle resampling is to consider it as an optimal transport (OT) problem~\cite{reich2013nonparametric}. Particularly, an entropy-regularised optimal transport-based resampler was proposed in~\cite{corenflos2021differentiable}. This OT-based resampler reformulates non-differentiable resampling steps as deterministic and differentiable operations by solving the optimal transport problem in resampling steps with the Sinkhorn algorithm~\cite{cuturi2013sinkhorn,feydy2019interpolating,peyre2019computational}. The OT-based resampler leads to an asymptotically consistent differentiable estimator of observation log-likelihoods, but it is biased due to the introduced non-zero regularisation term. Neural networks were utilised in the particle transformer proposed in~\cite{zhu2020towards}, which is designed based on the set transformer architecture~\cite{lee2019set,vaswani2017attention}. The particle transformer is permutation invariant and scale equivariant by design. The output of the particle transformer is a set of new particles with equal weights. The particle transformer is trained by maximising the likelihood of particles before resampling in a Gaussian mixture distribution consisting of Gaussian distributions with new particles as their means. However, a particle transformer trained on one task does not adapt to other tasks, implying that one needs to train a new particle transformer from scratch each time there is a new task.
	
	There are different strategies to train differentiable particle filters, which can be grouped into end-to-end training and those that involve individual training. In~\cite{jonschkowski18,karkus2018particle,wen2021end,ma2020particle,ma2019discriminative,chen2021deep}, differentiable particle filters are trained in an end-to-end manner where all components of differentiable particle filters are jointly trained by minimising an overall loss function via gradient descent. In contrast, in~\cite{jonschkowski18,karkus2021differentiable,karkus2019differentiable,lee2020multimodal}, it was proposed to first perform pre-training for independent modules of differentiable particle filters, then fine-tune these models jointly towards a task-specific target. 
	
	To optimise differentiable particle filters' parameters via gradient descent, a loss function or a training objective is needed. There are two major classes of loss functions that are often employed in training differentiable particle filters. The first class is the likelihood-based loss functions. In~\cite{naesseth2018variational,le2018auto,maddison2017filtering}, an evidence lower bound (ELBO) of observation log-likelihood was derived for particle filters. The particle filter-based ELBO is a surrogate objective to the marginal log-likelihood and can be used to learn both model and proposal parameters~\cite{corenflos2021differentiable,le2018auto}. Besides, a pseudo-likelihood loss was proposed in~\cite{wen2021end} to enable semi-supervised learning for differentiable particle filters when part of the ground-truth latent states is not provided. Another type of loss functions involves task-specific objectives. For example, in~\cite{jonschkowski18,karkus2018particle,wen2021end,corenflos2021differentiable,ma2020particle}, differentiable particle filters are used in visual robot localisation tasks so the prediction error between estimates and the ground-truth latent states (RMSE or the log density of ground-truth latent states) are included in their loss functions. In~\cite{chen2021deep}, the state space was discretised and Gaussian blur was applied to the ground-truth latent states in order to minimise the KL divergence between the ground-truth posteriors and the particle estimate of posteriors. In~\cite{dupty2021pf}, differentiable particle filters were used to guide the sampling process of the individualisation and refinement (IR) in graph neural networks (GNN), and the differentiable particle filters are learned by minimising the expected colouring error. A discriminative particle filter reinforcement learning (DPFRL) method was proposed in~\cite{ma2019discriminative}, where a differentiable particle filter is jointly trained end-to-end with a value network and a policy network by minimising a task-specific reinforcement learning loss. It was reported in~\cite{ma2020particle} that combining task-specific loss with negative log-likelihood loss gives the best empirical performance in numerical simulations.
	
	Having described the overall structure and design choices of differentiable particle filters, we now provide more detailed discussions on the key components of differentiable particle filters in the following sections.

	\section{Sampling Distributions of Differentiable Particle Filters}
	\label{sec:sampling_distribution}
	In particle filters, sampling distributions are the distributions from which particles are generated. The sampling distribution can be identical to the dynamic model $p(x_{t}| x_{t-1};\theta)$ (bootstrap particle filters), or designed as a proposal distribution $q(x_{t}| y_{t},x_{t-1};\phi)$ (guided particle filters) conditioned on both the state $x_{t-1}$ and the observation $y_t$. One of the key challenges in designing particle filters is the construction of sampling distributions that are easy to sample from, close to posterior distributions, and admit efficient weight update steps. We summarise below different design choices of sampling distributions in differentiable particle filters.
	% As discussed in Section~\ref{subsec:intro_dpfs}, both the state transition model $p(x_{t}| x_{t-1};\theta)$ and the proposal distribution $q(x_{t}| y_{t},x_{t-1};\phi)$ can be modelled with neural networks in differentiable particle filters. However, several differentiable particle filters construct their dynamic models as $p(x_{t}| y_t, x_{t-1};\theta)$~\cite{ma2020particle,karkus2021differentiable}, which do not fit into state-space models' assumptions. Nonetheless, this strategy has been successfully applied in a wide range of applications including robot localisation and sequence prediction tasks~\cite{ma2020particle,karkus2021differentiable}. Since in these differentiable particle filters the state transition model $p(x_{t}| y_t, x_{t-1};\theta)$ is conditioned on $y_t$ and $x_{t-1}$, which is similar to proposal distributions $q(x_{t}| y_{t},x_{t-1};\phi)$, we summarise different design choices of dynamic models and proposal distributions in this subsection.
	
	\subsection{Gaussian dynamic models}
	
	In several differentiable particle filter approaches~\cite{jonschkowski18,karkus2018particle,wen2021end,chen2022conditional,corenflos2021differentiable,chen2021deep,lee2020multimodal}, the transition of latent states from $x_{t-1}$ to $x_t$ is modelled by a deterministic function $g_\theta(\cdot): \mathcal{X}\times \mathbb{R}^{d_\mathcal{X}}\rightarrow \mathcal{X}$ with the previous state $x_{t-1}$ and a Gaussian noise term $\alpha_t$ as inputs and the new state $x_t$ as outputs. We name such dynamic models as Gaussian dynamic models. Generally speaking, Gaussian dynamic models used in differentiable particle filters can be described as follows:
	\begin{gather}
		x_t^i\sim\mathcal{N}(\mu_\theta(x_{t-1}^i), \Sigma_{d_\mathcal{X}})\,,\\
		x_t^i=g_\theta(x_{t-1}^i, \alpha_t^i)=\mu_\theta(x_{t-1}^i)+\alpha_t^i\,,
		\label{eq:gdm_general}
	\end{gather}
	where $\alpha_{t}^i \sim\mathcal{N}(\textbf{0}, \Sigma_{d_\mathcal{X}})$, and $\mu_\theta(\cdot):\mathcal{X}\rightarrow\mathbb{R}^{d_\mathcal{X}}$ is a differentiable function of $x_{t-1}^i$~\cite{jonschkowski18,karkus2018particle,wen2021end,corenflos2021differentiable,kloss2021train}.
	The covariance matrix $\Sigma_{d_\mathcal{X}}$ can be either designed manually~\cite{jonschkowski18,karkus2018particle,wen2021end,corenflos2021differentiable} or parameterised and learned from data~\cite{kloss2021train}. For data-adaptive covariance matrix $\Sigma_{d_\mathcal{X}}$, two different ways to learn the covariance matrix $\Sigma_{d_\mathcal{X}}$ were proposed in~\cite{kloss2021train}, the constant noise model and the heteroscedastic noise model. In constant noise models, the covariance matrix $\Sigma_{d_\mathcal{X}}=\text{diag}(\sigma_1^2, \cdots, \sigma_{d_\mathcal{X}}^2)$ is a diagonal matrix and keeps the same among different time steps. The diagonal elements $(\sigma_1^2, \cdots, \sigma_{d_\mathcal{X}}^2)$ of the covariance matrix $\Sigma_{d_\mathcal{X}}$ are set as learnable parameters and updated using gradient descent. For heteroscedastic noise models, the covariance matrix $\Sigma_{d_\mathcal{X}}$ is also a diagonal matrix, but it is predicted according to state values in heteroscedastic noise models. Specifically, for the transition from time step $t-1$ to time step $t$, a covariance matrix is computed for each particle:
	\begin{gather}
		(\sigma_1^i, \cdots, \sigma_{d_\mathcal{X}}^i)={\gamma_\theta(x_{t-1}^i)}\,,\\
		\Sigma_{d_\mathcal{X}}(x_{t-1}^i)=\text{diag}\Big((\sigma_1^i)^2, \cdots, (\sigma_{d_\mathcal{X}}^i)^2\Big)\,,
	\end{gather}
	where $\gamma_\theta(\cdot):\mathcal{X}\rightarrow\mathbb{R}^{d_\mathcal{X}}$ is a neural network. The noise term  $\alpha_t^i$ is sampled from $\mathcal{N}(\textbf{0}, \Sigma_{d_\mathcal{X}})$, where the covariance matrix $\Sigma_{d_\mathcal{X}}=\sum_{i=1}^{N_p}W_t^i\Sigma_{d_\mathcal{X}}(x_{t-1}^i)$ is computed as the weighted sum of individual covariance matrices $\Sigma_{d_\mathcal{X}}(x_{t-1}^i)$, with $W_t^i$ the normalised importance weight of $x_{t-1}^i$. With this setup, the covariance matrix $\Sigma_{d_\mathcal{X}}$ changes across time steps and hence the name heteroscedastic noise models.

	Here we only introduced Gaussian dynamic models with additive Gaussian noise which are most commonly used in differentiable particle filters~\cite{jonschkowski18,karkus2018particle,wen2021end,corenflos2021differentiable,kloss2021train}. One can also use multiplicative Gaussian noise and other reparameterisable distributions depending on applications~\cite{kingma2014auto,naesseth2018variational}.
	
	\vspace{1em}
	
	\subsection{Dynamic models constructed with normalising flows}
	
	A limitation of Gaussian dynamic models is their inability to adapt to complex systems where the latent state evolves according to complicated non-Gaussian dynamics.  To address this, it was proposed in~\cite{chen2021differentiable} to construct dynamic models in differentiable particle filters using normalising flows. Normalising flows are a family of invertible transformations that provides a flexible mechanism for constructing complex probability distributions~\cite{papamakarios2021normalizing}. By applying a series of invertible mappings on samples drawn from simple distributions, e.g. uniform or Gaussian distributions, normalising flows are able to transform simple distributions into arbitrarily complex distributions under some mild conditions~\cite{papamakarios2021normalizing}. In particular, it was proposed in~\cite{chen2021differentiable} to build flexible dynamic models ${p}(x_t | x_{t-1};\theta)$ upon base dynamic models $\tilde{p}(\tilde{x}_t | x_{t-1};\theta)$, e.g. Gaussian dynamic models. Denote by $\{\tilde{x}_t^i\}_{i=1}^{N_p}$ a set of particles sampled from $\tilde{p}(\tilde{x}_t | x_{t-1};\theta)$, in~\cite{chen2021differentiable} the particles $\tilde{x}_t^i$, $i \in \{1, \cdots, N_p\}$, are transformed by a normalising flow $\mathcal{T}_\theta(\cdot):\mathcal{X}\rightarrow \mathcal{X}$:
	\begin{gather}
            \label{eq:dynamic_model}
		{x}_t^i=\mathcal{T}_\theta(\tilde{x}_t^i)\sim p(x_t | x_{t-1}^i;\theta)\,,\; \text{where}\;\tilde{x}_t^i\sim \tilde{p}(\tilde{x}_t | x_{t-1}^i;\theta)\,.
	\end{gather}
	The probability density function $p(x_t | x_{t-1}^i;\theta)$ evaluated at $x_t^i $ can be computed through the change of variable formula:
	\begin{equation}
		\label{eq:density_nf}
		p(x_t^i | x_{t-1}^i;\theta)=\frac{\tilde{p}(\tilde{x}_t^i | x_{t-1}^i;\theta)}{\Big|\operatorname{det}J_{\mathcal{T_\theta}}(\tilde{x}_t^i)\Big|}\,,\; \text{where}\; \tilde{x}_t^i=\mathcal{T}^{-1}_\theta ({x}_t^i)\sim \tilde{p}(\tilde{x}_t | x_{t-1}^i;\theta)\,,
	\end{equation}
	where $\operatorname{det}J_{\mathcal{T_\theta}}(\tilde{x}_t^i)$ refers to the determinant of the Jacobian matrix $J_{\mathcal{T}_\theta}(\tilde{x}_t^i)=\frac{\partial \mathcal{T}_\theta(\tilde{x}_t^i)}{\partial \tilde{x}_t^i}$ evaluated at $\tilde{x}_t^i$.
	
	Because normalising flows can model complex non-Gaussian distributions, dynamic models constructed with normalising flows are theoretically more flexible than Gaussian dynamic models and require less prior knowledge of the system. In addition, experimental results reported in~\cite{chen2021differentiable} provided empirical evidence that using normalising flows-based dynamic models in differentiable particle filters can produce better object tracking performance than using Gaussian dynamic models. 
 
    % Other applications of normalising flows in sequential Monte Carlo methods can be found in~\cite{arbel2021annealed,matthews2022continual}, where normalising flows are employed to transport samples from the current posterior distribution to the next posterior distribution. However, the focus of~\cite{arbel2021annealed,matthews2022continual} is sequential Monte Carlo samplers and it was assumed in~\cite{arbel2021annealed,matthews2022continual} that the posterior distributions can be evaluated point-wise up to a normalising constant, which distinguishes them from the problem setup in differentiable particle filters, so we choose not to present more details on this direction in this paper.
	\subsection{Dynamic models conditioned on observations}
	%	 Secondly, in some applications the actual dynamics of latent states might be more complicated than an unimodal Gaussian.
	In several differentiable particle filtering approaches, dynamic models are constructed as $p(x_{t}| y_{1:t}, x_{t-1};\theta)$ and used to generate new particles~\cite{ma2020particle,karkus2021differentiable}. This strategy has been successfully applied in a wide range of applications including robot localisation and sequence prediction tasks~\cite{ma2020particle,karkus2021differentiable}. For instance, the dynamic model of the differentiable particle filter proposed in~\cite{karkus2021differentiable} employs a neural network to model the displacement of robots. The neural network takes observation images $(y_{t-1},y_{t})$ as its inputs, and outputs the mean and the variance of a Gaussian distribution that models the distribution of the robot's displacement. New particles $\{x_t^i\}_{i=1}^{N_p}$ are obtained by adding Gaussian noise drawn from the learned Gaussian distribution to $\{x_{t-1}^i\}_{i=1}^{N_p}$. A similar dynamic model is used in~\cite{ma2019discriminative} to assist the training of a reinforcement learning algorithm.
	In~\cite{ma2020particle}, recurrent neural networks (RNNs) are used to construct dynamic models in differentiable particle filters. Specifically, it was proposed in~\cite{ma2020particle} to model the evolution of the latent state using LSTM networks~\cite{hochreiter1997long} or GRU networks~\cite{cho2014properties}, hence the name particle filter recurrent neural networks (PFRNNs). In the PFRNN method, particles at the $t$-th time step are generated through $x_t^i=\text{RNN}_\theta(x_{t-1}^i, y_t, \beta_t^i)$. $\beta_t^i$ is sampled from a Gaussian distribution with learnable mean and variance which are the output of a function of $x_{t-1}^i$ and $y_t$. Note that the latent state $x_t$ in the PFRNN may not be the state of direct interest, and the weighted average of particles $\bar{x}_t=\sum_i^{N_p}W_t^i x_t^i$ is taken as the input of a task-dependent function to predict the state of interest. For example, in the robot localisation experiment presented in~\cite{ma2020particle}, the task-dependent function maps $\bar{x}_t$ to robot locations, and in classification tasks, the outputs of the task-dependent function are probability vectors of categorical distributions.

	\subsection{Proposal distributions}
	
	To the best of our knowledge, the majority of existing differentiable particle filter approaches are restricted to the bootstrap particle filtering framework which does not utilise observations to construct proposal distributions. When latent states evolve according to a simple dynamics and the relationship between $x_t$ and $y_t$ can be easily speculated with prior knowledge, optimal proposals can be explicitly derived or approximated with closed-form expressions~\cite{chopin2020introduction}. 
	However, in more complex environments such as vision-based robot localisation, it is intractable to explicitly derive or approximate optimal proposals and therefore one needs to learn proposal distributions from data for differentiable particle filters. 
 
    Different ways to learn proposal distributions in differentiable particle filters have been proposed~\cite{chen2021differentiable,gama2022unrolling,gama2023unsupervised}. In~\cite{gama2022unrolling,gama2023unsupervised}, the proposal distribution is constructed as a Gaussian distribution whose mean and covariance matrix are given by a neural network with the estimate of the previous latent state and the current observation as inputs. To provide a mechanism for constructing proposal distributions more flexible than Gaussian distributions, it was proposed in~\cite{chen2021differentiable} to construct proposal distributions for differentiable particle filters with conditional normalising flows~\cite{winkler2019learning}. Specifically, proposal distributions in~\cite{chen2021differentiable} are constructed by stacking a conditional normalising flow $\mathcal{G}_\phi(\cdot):\mathcal{X}\times\mathcal{Y}\rightarrow\mathcal{X}$ upon the dynamic model:
	\begin{gather}
        \label{eq:proposal_cnf}
		x_t^i=\mathcal{G}_\phi(\check{x}_t^i, y_t)\sim q(x_{t}| y_{t},x_{t-1}^i;\phi)\,,\; \text{where}\; \check{x}_t^i\sim p(\check{x}_t | x_{t-1}^i;\theta)\,.
	\end{gather}
	The conditional normalising flow $\mathcal{G}_\phi(\cdot)$ is an invertible function of the particle $\check{x}_t^i\sim p(\check{x}_t | x_{t-1};\theta)$ when the observation $y_t$ is given. Since information from the latest observation is taken into account, the conditional normalising flow $\mathcal{G}_\phi(\cdot)$ has the potential to migrate samples drawn from the prior to regions that are closer to the posterior. Similar to Eq.~(\ref{eq:density_nf}), the proposal density can be evaluated by applying the change of variable formula:
	\begin{gather}
            \label{eq:cnf_proposal_density}
		q(x_{t}^i| y_{t}, x_{t-1}^i;\phi)=\frac{p\left(\check{x}_{t}^{i} | x_{t-1}^{i}; \theta\right)}{\Big|\operatorname{det}J_{\mathcal{G}_{\phi}}\left(\check{x}_{t}^{i}\right)\Big|}\,,\; \text{where}\; \check{x}_{t}^{i}=\mathcal{G}_\phi^{-1}(x_t^i, y_t)\sim p(\check{x}_t | x_{t-1}^i;\theta)\,.
	\end{gather}
 $\operatorname{det}J_{\mathcal{G_\phi}}(\check{x}_{t}^{i})$ refers to the determinant of the Jacobian matrix, i.e., $J_{\mathcal{G}_\phi}(\check{x}_t^i)=\frac{\partial \mathcal{G}_\phi(\check{x}_t^i, y_t)}{\partial \check{x}_t^i}$ evaluated at $\check{x}_t^i$.
	
	\section{Measurement Models of Differentiable Particle Filters}
	\label{sec:measurement_model}
	In this section we briefly review different designs of measurement models in the differentiable particle filter literature. We group measurement models used in differentiable particle filters into four categories. The first category estimates $p(y_t| x_t;\theta)$ using known distributions with unknown parameters~\cite{corenflos2021differentiable,naesseth2018variational}, e.g. parameterised Gaussian distributions or Poisson distributions. The second category relies on neural networks to learn a scalar function to approximate $p(y_t| x_t;\theta)$~\cite{jonschkowski18,karkus2018particle}. The third category measures the discrepancy or similarity between observations and latent states in a learned feature space~\cite{wen2021end,chen2021differentiable}. The last category utilises conditional normalising flows to learn the generating process of $y_t$ and consequently the density of $y_t$ given $x_t$~\cite{chen2022conditional}. More details on these different types of measurement models are provided below.
	
	\subsection{Measurement models built with known distributions}
	
	We first introduce measurement models used in differentiable particle filters that capture the relationship between $x_t$ and $y_t$ using known distributions with unknown parameters. An example of this type of measurement model can be found in~\cite{le2018auto}, where differentiable particle filters are used to learn the parameters of a linear-Gaussian state-space model:
	\begin{align}
		x_0 &\sim\mathcal{N}({0, 1})\,, \\
		x_t|x_{t-1} &\sim\mathcal{N}(\theta_{1} x_{t-1}, 1)\,,\; \text{for}\; t\geq 1\,, \\
		y_t|x_t &\sim\mathcal{N}(\theta_{2} x_{t}, {0.1})\,,\; \text{for}\; t\geq 0\,,
	\end{align}
	where $\theta=\{\theta_1, \theta_2\}$ are parameters to be estimated. In the above example, the likelihood of $y_t$ given $x_t$ is established as the probability density function of a Gaussian distribution with learnable parameters.
	
	Measurement models built with known distributions are limited to cases where the relationship between $y_t$ and $x_t$ can be explicitly inferred. In more complex systems where observations are high-dimensional unstructured data like images, one often needs to design measurement models using more powerful tools such as neural networks.
	
	\subsection{Measurement models as scalar functions built with neural networks}
	\label{subsec:measurement_nn}
	In scenarios where it is non-trivial to model $p(y_t | x_t ;\theta)$ with explicit closed-form expressions, several differentiable particle filtering approaches employ neural networks to approximate $p(y_t | x_t ;\theta)$~\cite{jonschkowski18,karkus2018particle}. A straightforward idea is to learn a scalar function $l_\theta(\cdot):\mathcal{Y}\times\mathcal{X}\rightarrow \mathbb{R}$ constructed by neural networks and consider its output as a measure of the compatibility between $x_t$ and $y_t$. In~\cite{jonschkowski18}, observations are first mapped into a feature space through $e_t=E_\theta(y_t)\in\mathbb{R}^{d_e}$, where $E_\theta(\cdot):\mathcal{Y}\rightarrow\mathbb{R}^{d_e}$ is a neural network. Then another neural network $h_\theta(\cdot):\mathbb{R}^{d_e}\times \mathcal{X}\rightarrow \mathbb{R}$ with $e_t$ and $x_t^i$ as inputs is employed, whose outputs are considered as the unnormalised likelihood of $y_t$ given $x_t^i$:
	\begin{align}
		e_t&=E_\theta(y_t)\,,\\
		p(y_t| x_t^i;\theta) &\propto l_\theta(y_t, x_t^i)\nonumber\\
		&=h_\theta\big(E_\theta(y_t), x_t^i\big)\,.
	\end{align}
	In~\cite{karkus2018particle}, differentiable particle filters are applied in robot localisation tasks where a global 2D map is provided and observations are 3D local images. For this environment, spatial transformer networks~\cite{jaderberg2015spatial} are used in~\cite{karkus2018particle} to generate local maps $M_{x_t^i}\in\mathbb{R}^{h}\times\mathbb{R}^{w}$ based on the global map $\mathbb{M}\in\mathbb{R}^{H}\times\mathbb{R}^{W}$ and particle locations $x_t^i$, where $h$, $w$, $H$, and $W$ are the height and width of the local and global maps, respectively. Then, the 2D local map $M_{x_t^i}$ is compared with the 3D local image $y_t$ through a neural network $h_\theta(\cdot):\mathcal{Y}\times(\mathbb{R}^{h}\times\mathbb{R}^{w})\rightarrow\mathbb{R}$:
	\begin{align}
		M_{x_t^i}&=\textbf{ST}_\theta(\mathbb{M}, x_t^i)\,,\\
		p(y_t| x_t^i;\theta) &\propto l_\theta(y_t, x_t^i)\,\\
		&=h_\theta(y_t, M_{x_t^i})\,,
	\end{align}
	where $\textbf{ST}_\theta(\cdot):(\mathbb{R}^{H}\times\mathbb{R}^{W})\times\mathcal{X}\rightarrow \mathbb{R}^{h}\times\mathbb{R}^{w}$ refers to a spatial transformer network~\cite{jaderberg2015spatial}.
 
 One open question for this category of measurements is how to normalise the learned unnormalised likelihood of $y_t$ given $x_t^i$. When the loss function used to train differentiable particle filters is maximum likelihood-based, a measurement model built with a neural network and scalar, unnormalised output may assign arbitrarily high likelihood to any input, leading to poor generalisation performance of differentiable particle filters with these setups.
	
	\subsection{Measurement models as feature similarities}
	
	Instead of solely relying on neural networks to build measurement models for differentiable particle filters, there are other differentiable particle filtering approaches that consider neural networks as feature extractors and employ user-defined metric functions to compare observation features and state features. In~\cite{wen2021end}, two neural networks $E_\theta(\cdot):\mathcal{Y}\rightarrow\mathbb{R}^{d_e}$ and $O_\theta(\cdot):\mathcal{X}\rightarrow\mathbb{R}^{d_e}$ are used to extract features from observations and states respectively, then $p(y_t| x_t^i;\theta)$ is assumed to be proportional to the reciprocal of the cosine distance between $E_\theta(y_t)$ and $O_\theta(x_t^i)$:
	\begin{gather}
		e_t=E_\theta(y_t)\,, o_t^i=O_\theta(x_t^i)\,,\\
		c(e_t, o_t^i)=1-\frac{e_t \cdot o_t^i}{||e_t||_2 || o_t^i||_2}\,,\\
		p(y_t| x_t^i;\theta) \propto \frac{1}{c(e_t, o_t^i)}\,,
	\end{gather}
	where $c(\cdot):\mathbb{R}^{d_e}\times \mathbb{R}^{d_e} \rightarrow \mathbb{R}$ refers to the cosine distance, $e_t \cdot o_t^i$ denotes the inner product of features vectors, and $||\cdot||_2$ is the $L_2$ norm of feature vectors. In the robot localisation experiment of~\cite{corenflos2021differentiable}, the relationship between observation features and state features is modelled as follows:
	\begin{gather}
		e_t=E_\theta(y_t)\sim \mathcal{N}(o_t^i,\;\sigma^2_\text{obs}\textbf{I})\,.
	\end{gather}
	In other words, the conditional likelihood of $y_t$ given $x_t$ is computed as the Gaussian density of the observation feature $e_t$ with $o_t^i=O_\theta(x_t^i)$ and $\sigma^2_\text{obs}\textbf{I}$ as its mean and variance.
	
	To ensure that the observation feature extractor $E_\theta(\cdot)$ maintains essential information from $y_t$, it was suggested to add an auto-encoder loss into the training objective~\cite{wen2021end,corenflos2021differentiable}:
	\begin{gather}
		\mathcal{L}_{\text{AE}}=\sum_{t=0}^{T} ||D_\theta(E_\theta(y_t))-y_t||_2^2\,,
	\end{gather}
	where $T$ is the length of sequences, $E_\theta(\cdot):\mathcal{Y}\rightarrow\mathbb{R}^{d_e}$ and $D_\theta(\cdot):\mathbb{R}^{d_e}\rightarrow\mathcal{Y}$ are the observation encoder and the observation decoder, respectively.
	
	\subsection{Measurement models built with conditional normalising flows}
	
	A conditional normalising flow-based measurement model was proposed for differentiable particle filters in~\cite{chen2022conditional}. Conditional normalising flow-based measurement models provide a mechanism to compute valid conditional probability density of $y_t$ given $x_t$. In particular, the measurement model proposed in~\cite{chen2022conditional} computes $p(y_t| x_t^i;\theta)$ via:
	\begin{gather}
		p(y_t| x_t^i;\theta)=p_z(\mathcal{F}_\theta(y_t, x_t^i))\bigg|\operatorname{det}\frac{\partial \mathcal{F}_\theta(y_t, x_t^i)}{\partial y_t}\bigg|\,,
		\label{eq:cnf_measurement}
	\end{gather}
	where $\mathcal{F}_\theta(\cdot):\mathcal{Y}\times\mathcal{X}\rightarrow\mathbb{R}^{d_\mathcal{Y}}$ is a parameterised conditional normalising flow, and $\bigg|\operatorname{det}\frac{\partial \mathcal{F}_\theta(y_t, x_t^i)}{\partial y_t}\bigg|$ is the determinant of the Jacobian matrix $\frac{\partial \mathcal{F}_\theta(y_t, x_t^i)}{\partial y_t}$ evaluated at $y_t$. The base distribution $p_z(\cdot)$ defined on $\mathbb{R}^{d_\mathcal{Y}}$ can be user-specified and is often chosen as a simple distribution such as an isotropic Gaussian.
	
	In scenarios where observations $y_t$ are high-dimensional data like images, the evaluation of Eq.~\eqref{eq:cnf_measurement} with raw observations can be computationally expensive. In addition, when observations contain too much irrelevant information, the construction of measurement models with raw observations even becomes more challenging. As an alternative solution, it was proposed in~\cite{chen2022conditional} to map observations $y_t$ to a lower-dimensional feature space through a neural network $E_\theta(\cdot):\mathcal{Y}\rightarrow\mathbb{R}^{d_e}$. With the assumption that the conditional probability density $p(e_t| x_t^i;\theta)$ of observation features $e_t=E_\theta(y_t)$ given state is an approximation of the actual measurement likelihood $p(y_t| x_t^i;\theta)$, in~\cite{chen2022conditional} the conditional likelihood of observations is computed by:
	\begin{align}
		p(y_t| x_t^i;\theta)&\approx p(e_t| x_t^i;\theta)\\
		&=p_z(\mathcal{F}_\theta(e_t, x_t^i))\bigg|\operatorname{det}\frac{\partial \mathcal{F}_\theta(e_t, x_t^i)}{\partial e_t}\bigg|\,.
	\end{align}
	
	\section{Differentiable Resampling}
	\label{sec:resampling}
	Resampling is the key operation that distinguishes sequential Monte Carlo (SMC) methods from sequential importance sampling (SIS). It is also a major obstacle to achieve fully differentiable particle filters. Due to the fact that the resampling output changes in a discrete manner, it has been widely documented that the resampling step is not differentiable~\cite{jonschkowski18,karkus2018particle,corenflos2021differentiable,rosato2022efficient}. Additionally, since the importance weights of particles are set to be a constant after resampling, the gradients of particle weights at the $t$-th time step w.r.t. particle weights at time step $t'< t$ are always zero, rendering the operation non-differentiable in a machine learning context~\cite{karkus2018particle}.
	
	To estimate the gradients of resampling steps, different solutions have been proposed in~\cite{karkus2018particle,corenflos2021differentiable,zhu2020towards} to achieve differentiable resampling. We review these techniques below.
	
	\subsection{Soft resampling}
 \label{subsec:soft_resampling}
	The soft resampling method proposed in~\cite{karkus2018particle} attempts to generate non-zero values for the gradients $\frac{\partial\tilde{w}_t^i}{\partial {w}_t^i}$, where $\partial\tilde{w}_t^i$ and $\partial {w}_t^i$ are unnormalised particle weights after and before resampling, respectively. Denote by $\{W_t^i\}_{i=1}^{N_p}$ the set of normalised particle weights before resampling, the soft resampling constructs an importance multinomial distribution $\textbf{Mult}(\tilde{W}_t^1, \cdots, \tilde{W}_t^{N_p})$ defined with $\{\tilde{W}_t^i\}_{i=1}^{N_p}$:
	\begin{gather}
            \label{eq:soft_weights}
		\tilde{W}_t^i=\lambda W_t^i+(1-\lambda)\frac{1}{N_p}\,.
 	\end{gather}
	The indices of selected particles are sampled from the importance multinomial distribution $\textbf{Mult}(\tilde{W}_t^1, \cdots, \tilde{W}_t^{N_p})$. The importance distribution $\textbf{Mult}(\tilde{W}_t^1, \cdots, \tilde{W}_t^{N_p})$ can be interpreted as a linear interpolation between the original multinomial distribution $\textbf{Mult}(W_t^1, \cdots, W_t^{N_p})$  and a multinomial distribution $\textbf{Mult}(\frac{1}{N_p},\cdots,\frac{1}{N_p})$ with equal weights $\frac{1}{N_p}$. To correct the bias caused by sampling from the importance distribution, soft resampling updates particle weights after resampling as:
	\begin{gather}
		\label{eq:soft_resampling_weights_corrected}
		\tilde{{w}}_t^i=\frac{W_t^i}{\tilde{W}_t^i}=\frac{W_t^i}{\lambda W_t^i+(1-\lambda){1}/{N_p}}\,.
	\end{gather}
	Since the modified weights in Eq.~\eqref{eq:soft_resampling_weights_corrected} is a function of the original weight $W_t^i=\frac{w_t^i}{\sum_{j=1}^{N_p}w_t^j}$, the gradient $\frac{\partial\tilde{w}_t^i}{\partial {w}_t^i}$ yields non-zero values with the soft resampling method.
	
	However, soft resampling still relies on sampling from multinomial distributions, thus it does not change the discrete nature of the resampling step. In other words, the soft resampling outputs may still change in a discrete manner with slight changes in $W_t^i$, i.e. the non-differentiable part of resampling is ignored in soft resampling~\cite{corenflos2021differentiable}. Nonetheless, soft resampling has been widely used in the differentiable particle filter literature, due to its simple implementation and competitive empirical performance in practice~\cite{wen2021end,ma2020particle,ma2019discriminative,karkus2021differentiable,karkus2019differentiable}.
	
	\subsection{Differentiable resampling via entropy-regularised optimal transport}
	
	An optimal transport (OT) map-based differentiable resampling algorithm was proposed in~\cite{corenflos2021differentiable}, which we refer to as OT-resampler. The OT-resampler finds the optimal transport map between an equally weighted empirical distribution $\rho(x_t)=\frac{1}{N_p}\sum_{i=1}^{N_p}\,\delta_{x_{t}^i}(x_{t})$ and the target empirical distribution $\hat{p}(x_t|y_{1:t};\theta)=\sum_{i=1}^{N_p}W_t^i\, \delta_{x_{t}^i}(x_{t})$. Since the original optimal transport problem is computationally expensive and not differentiable, the optimal transport map is obtained by solving an entropy-regularised version of optimal transport problems~\cite{villani2003topics,villani2008optimal} with the Sinkhorn algorithm~\cite{cuturi2013sinkhorn,feydy2019interpolating,peyre2019computational} which is efficient and differentiable.
	
	% 	, where $\xi$ is the regularization coefficient in the entropy-regularised OT problem~\cite{peyre2019computational}
	
	Let us first consider the application of optimal transport in particle resampling without regularisation. Ideally, we want to find a transportation map $T: \mathcal{X}\rightarrow\mathcal{X}$ such that the resulting distribution $T\#\rho(x_t)=\frac{1}{N_p}\sum_{i=1}^{N_p}\,\delta_{\tilde{x}_t^i}(x_{t})$ with $\tilde{x}_t^i=T(x_t^i)$ exactly matches the target distribution $\hat{p}(x_t|y_{1:t};\theta)=\sum_{i=1}^{N_p}W_t^i\, \delta_{x_{t}^i}(x_{t})$. However, this is not possible because the resulting distribution has equal weights while the target distribution does not. An approximation $\hat{T}:x_t^i\in\mathcal{X}\rightarrow N_p\sum_{j=1}^{N_p} P^\text{OT}(i,j)x_t^j \in\mathcal{X}$ of the optimal transportation map $T$ is the so-called barycentric projection~\cite{corenflos2021differentiable,peyre2019computational}:
	\begin{gather}
		\hat{x}_t^i=N_p\sum_{j=1}^{N_p} P^\text{OT}(i,j)x_t^j\,,\\
		\label{eq:barycentric}
		\hat{X}_t=N_p P^{\text{OT}}X_t\,,\;\text{where}\; P^{\text{OT}}\in\mathbb{R}^{N_p\times N_p}\,.
	\end{gather}
	 The matrix $P^{\text{OT}}$ is the optimal coupling that produces the minimal transportation cost $\sum_{i,j}^{N_p}P^{\text{OT}}(i,j)\mathcal{D}(x_t^i,x_t^j)$ for a predefined distance metric $\mathcal{D}(\cdot, \cdot):\mathcal{X}\times\mathcal{X}\rightarrow\mathbb{R}$, and satisfies $P^{\text{OT}}\textbf{1}_{N_p}=(\frac{1}{N_p}, \cdots, \frac{1}{N_p})^\top$, $(P^{\text{OT}})^\top\textbf{1}_{N_p}=(W_t^1, \cdots, W_t^{N_p})^\top$. We use $P^{\text{OT}}(i,j)$ to denote the $(i,j)$-th element of $P^{\text{OT}}$ and $\textbf{1}_{N_p}\in \mathbb{R}^{N_p}$ a $N_p$-dimensional column vector with all entries set to 1. In Eq.~\eqref{eq:barycentric}, the $i$-th row of $\hat{X}_t\in\mathbb{R}^{N_p\times d_\mathcal{X}}$ and ${X}_t\in\mathbb{R}^{N_p\times d_\mathcal{X}}$ are the $i$-th transported particle $\hat{x}_t^i$ and the $i$-th original particle ${x}_t^i$, respectively. The transported particle $\hat{x}_t^i$ is the barycentre of the empirical distribution $\tau_t^i(x_t)={N_p}\sum_{j=1}^{N_p}{P^\text{OT}(i,j)}\delta_{x_t^j}(x_t)$, hence the name barycentric projection. The distribution $\hat{T}\#\rho(x_t)=\frac{1}{N_p}\sum_{i=1}^{N_p}\,\delta_{\hat{x}_t^i}(x_{t})$ is considered as the resampling output of the OT-resampler.
	
	%	and leads to unbiased estimates of $\sum_{i=1}^{N_p}W_t^i f(x_t^i)$ for affine functions $f(\cdot):\mathcal{X}\rightarrow\mathbb{R}$~\cite{reich2013nonparametric}.
	
	However, as aforementioned, the original OT problem leads to high computational overhead and is not differentiable. In~\cite{corenflos2021differentiable}, to make the OT-based resampling efficient and differentiable, it was proposed to solve the dual form of an entropy-regularised version of the original OT problem with the Sinkhorn algorithm~\cite{cuturi2013sinkhorn}. Using $P_\xi^{\text{OT}}\in\mathbb{R}^{N_p\times N_p}$ to refer to the entropy-regularised coupling with regularisation coefficient $\xi$, resampled particles are obtained by applying the corresponding barycentric projection $\hat{T}_\xi:x_t^i\in\mathcal{X}\rightarrow N_p\sum_{j=1}^{N_p} P_\xi^\text{OT}(i,j)x_t^j \in\mathcal{X}$:
	\begin{gather}
		\hat{x}_{t,\xi}^i=N_p\sum_{j=1}^{N_p} P_\xi^\text{OT}(i,j)x_t^j\,,\\
		\label{eq:barycentric_regularised}
		\hat{X}_{t,\xi}=N_p P_\xi^{\text{OT}}X_t\,.
	\end{gather}
	% Denote by $P^{\text{OT}}(i,j)$ the element at the $i$-th row and the $j$-th column of $P^{\text{OT}}$ and $\textbf{1}_{N_p}\in \mathbb{R}^{N_p}$ a $N_p$-dimensional vector with all entries set to 1, the matrix $P^{\text{OT}}$ yields a joint distribution $\{P^{\text{OT}}(i,j),(x_t^i,x_t^j)\}_{i,j=1}^{N_p}$ of discrete distributions $\{\frac{1}{N_p}, {x}_t^i\}_{i=1}^{N_p}$ and $\{W_t^j, x_t^j\}_{j=1}^{N_p}$, i.e. $P^{\text{OT}}\textbf{1}_{N_p}=(\frac{1}{N_p}, \cdots, \frac{1}{N_p})^\top$ and $(P^{\text{OT}})^\top\textbf{1}_{N_p}=(W_t^1, \cdots, W_t^{N_p})^\top$.	
	While the OT-resampler is fully differentiable, the regularisation coefficient $\xi$ introduces biases into the OT-resampler, and the bias only vanishes when $\xi\rightarrow 0$. Despite the biasedness of the OT-resampler, the OT-resampler has been shown to outperform soft resampling in experiments conducted in~\cite{corenflos2021differentiable}.
	
	\subsection{Differentiable resampling via neural networks}
	
    The particle transformer proposed in~\cite{zhu2020towards} is a neural network architecture designed for performing differentiable particle resampling. Since the inputs of the resampling step are point sets, the particle transformer is designed to be permutation-invariant and scale-equivariant by utilising the set transformer~\cite{lee2019set}. The input of the particle transformer is a set of weighted particles $\{W_t^i, x_t^i\}_{i=1}^{N_p}$, and the output of the particle transformer is a set of new particles $\{\tilde{x}^i_t\}_{i=1}^{N_p}$:
	\begin{gather}
		\{\tilde{x}^i_t\}_{i=1}^{N_p}=\textbf{PT}_\theta\big(\{W_t^i, x_t^i\}_{i=1}^{N_p}\big)\,,
	\end{gather}
	where $\textbf{PT}_\theta(\cdot)$ denotes the particle transformer. The output particles are assigned with equal weights $\tilde{W}_t^i=\frac{1}{N_p}$.
	
	Since it is a neural network-based resampler, a training objective is needed to train the particle transformer. In~\cite{zhu2020towards}, the particle transformer is optimised by maximising the likelihood of original samples $\{{x}^i_t\}_{i=1}^{N_p}$ in the distribution after resampling, and the likelihood of a particle $x_t^i$ is weighted by its importance weight $W_t^i$. To convert empirical distributions into continuous distributions with proper probability density functions, it was proposed in~\cite{zhu2020towards} to construct a Gaussian mixture distribution based on the empirical distribution after resampling (one Gaussian kernel per particle). Components of the Gaussian mixture model are weighted equally, since particles after resampling have uniform weights. The loss function used to train the particle transformer is defined as the weighted likelihood of the original samples in the Gaussian mixture distribution:
	\begin{equation}
		\label{eq:loss_transformer}
		-\sum_{i=1}^{N_p}W_t^i\log\; \sum_{j=1}^{N_p}\frac{1}{\sqrt{|\Sigma|}}\exp\big(-\frac{1}{2}(x_t^i-\tilde{x}_t^j)^\top\Sigma^{-1}(x_t^i-\tilde{x}_t^j)\big)\,,
	\end{equation}
	where $\Sigma$ is a specified covariance matrix of the Gaussian mixture model, and $|\Sigma|$ refers to the determinant of $\Sigma$. After training the particle transformer individually, the particle transformer was integrated into differentiable particle filters and trained jointly by minimising a task-specific loss function~\cite{zhu2020towards} .
	
Particle transformer requires pre-training, implying that one needs to collect training data beforehand, which may not be practical in some applications. In addition, a particle transformer trained for a specific task may not fit into other tasks and requires re-training to deploy it in a new task. Additionally, experimental results presented in~\cite{zhu2020towards} show that the best performance is achieved if one does not backpropagate through the particle transformer when training differentiable particle filters in the reported experiment.
	
	\section{Loss Functions for Training Differentiable Particle Filters}
	\label{sec:loss_function}
	With all the components we have described, we need a loss function to train neural networks used to build the above key components of differentiable particle filters. In the differentiable particle filter literature, loss functions can be mainly grouped into three categories, supervised losses, semi-supervised losses, and variational objectives, as described below.

	\vspace{1em}
	
	\subsection{Supervised losses}
	
	Supervised losses are a popular choice in differentiable particle filters when ground-truth latent states are available~\cite{jonschkowski18,karkus2018particle,chen2021differentiable,chen2022conditional,corenflos2021differentiable}, which we use $x_t^*$ to denote the ground-truth latent states. Training differentiable particle filters with supervised losses implies the ground-truth latent state  $x_t^*$ is available, therefore we can evaluate the similarity between the estimated state $\bar{x}_t$ given by differentiable particle filters and the ground-truth $x_t^*$. There are two major classes of supervised losses used in differentiable particle filters, i.e., the root mean square error (RMSE) and the negative (state) likelihood loss.
	
	We first introduce the RMSE loss. The RMSE loss computes the distance between the predicted states $\bar{x}_t$ and the ground-truth latent states $x_t^*$:
	\begin{gather}
		\bar{x}_t=\sum_{i=1}^{N_p}W_t^i x_t^i\,,\\
		\mathcal{L}_{\text{RMSE}}:=\sqrt{\frac{1}{T+1}\sum_{t=0}^{T}||x_t^*-\bar{x}_t||_2^2}\,,
	\end{gather}
	where $||\cdot||_2$ denotes the $L_2$ norm, and $T+1$ is the total number of time steps.
	Minimising the RMSE loss is equivalent to maximising the log-likelihood of the ground-truth latent state $x_t^*$ in the Gaussian distribution with mean $\bar{x}_t$ and an identity covariance matrix. 
	
	In contrast to the RMSE loss which considers only the final estimates, the so-called data likelihood loss takes into account all the particles~\cite{jonschkowski18}. The log-likelihood loss $\mathcal{L}_{\text{LL}}$ computes the logarithm of probability density of ground-truth latent states $x_t^*$ in a Gaussian mixture distribution defined by a set of weighted particles $\{W_t^i, {x}_t^i\}_{i=1}^{N_p}$:
	\begin{gather}
		\mathcal{L}_{\text{LL}}:=-\frac{1}{T+1}\sum_{t=0}^{T} \log\sum_{i=1}^{N_p}\frac{W_t^i}{\sqrt{|\Sigma|}}\exp\big(-\frac{1}{2}(x^*_t-{x}_t^i)^\top\Sigma^{-1}(x_t^*-x_t^i)\big)\,,
	\end{gather}
	where $\Sigma$ is a specified covariance matrix of the Gaussian mixture model, and $|\Sigma|$ refers to the determinant of $\Sigma$.
	
	Despite that supervised losses have been empirically proven to be effective among various differentiable particle filtering methods~\cite{jonschkowski18,karkus2018particle,chen2021differentiable,chen2022conditional,corenflos2021differentiable}, one limitation of supervised losses targeting either the posterior mean or the largest posterior mode is that they may hinder the particle filter's ability to capture less prominent modes or the tail of the distribution.
    % To better model posterior distributions with a mixture of different modes, a regime-switching mechanism was introduced for DPFs in~\cite{li2023differentiable} by learning a set of unknown candidate models. It was assumed in~\cite{li2023differentiable} that the regime-switching dynamic, i.e. the transition probability between different modes, is known, therefore it remains an open question how to capture the multi-modality of posteriors if one does not know the interplay between different modes in the system of interest. 
 
    Another limitation of training differentiable particle filters with supervised losses is that they require sufficient training data with ground-truth latent state information, which can be expensive to collect and are often unavailable in real-world applications. We discuss alternative choices of training loss functions below.
	
	\subsection{Semi-supervised loss}
	
	When ground-truth latent states are not provided, a common choice of training objectives for differentiable particle filters is the marginal likelihood of observations $p(y_{1:t};\theta)$. In~\cite{andrieu2004particle}, it was proposed to search for point estimates of static parameters in particle filters by maximising the log-likelihood $L_t(\theta)=\log p(y_{1:t};\theta)$ using noisy gradient algorithms. However, this approach requires the estimation of sufficient statistics of joint distributions whose dimension increases over time. An alternative is to use a pseudo-likelihood~\cite{andrieu2005line}. Specifically, it was proposed in~\cite{andrieu2005line} to split the data into slices and consider the product of each block's likelihood as the pseudo-likelihood of the whole trajectory. Inspired by~\cite{andrieu2005line}, a pseudo-likelihood loss was introduced in~\cite{wen2021end} to enable semi-supervised learning in differentiable particle filters.
	
	To introduce the semi-supervised loss, one first derives a logarithmic pseudo-likelihood $\mathcal{Q}(\theta)$. Consider $m$ slices $X_b:=x_{bL:(b+1)L-1}$ and $Y_b:=y_{bL:(b+1)L-1}$ of $\{x_t\}_{t=0}^{T}$ and $\{y_t\}_{t=1}^{T}$, where $L$ is the time length of each block, $X_b$ and $Y_b$ are the $b$-th block of states and observations, respectively. The log pseudo-likelihood for $m$ blocks of observations is defined as:
	\begin{gather}
		{\mathcal{Q}}(\theta):=\sum_{b=0}^{m-1}\log p(Y_b ;\theta)\,.
	\end{gather}
	Compared to the true log-likelihood $L_t(\theta)=\log p(y_{1:t};\theta)$, the pseudo-likelihood overcomes the increasing dimension problem by ignoring the dependency between blocks.

	At the $b$-th block, denote by $\theta_b$ the current estimate of parameter value and $\theta^*$ the optimal parameter value, $Q(\theta, \theta_b)$ is maximised over $\theta$, which is equivalent to maximising the pseudo-likelihood ${\mathcal{Q}}(\theta)$, where $Q(\theta, \theta_b)$ is defined as follows:
	\begin{align}
		Q(\theta, \theta_b):=\int_{\mathcal{X}^L\times\mathcal{Y}^L}\log( p(X_b, Y_b;\theta))\cdot p(X_b| Y_b;\theta_b)p(Y_b;\theta^*) \text{d}X_b\text{d}Y_b\,.
	\end{align}
	Particularly, $Q(\theta, \theta_b)$ is maximised through gradient descent with the initial value of $\theta$ set to be $\theta_b$. 
	With the pseudo-likelihood objective, we can train differentiable particle filters in a semi-supervised manner as follows:
	\begin{align}
		\theta=\underset{\theta\in\Theta}{\operatorname{argmin}}\;\bigg(\lambda_1 \mathcal{L}(\theta)-\frac{\lambda_2}{m}{Q}(\theta, \theta_b)\bigg)\,,\; \text{if ground-truth latent states are available,}\\
		\theta=\underset{\theta\in\Theta}{\operatorname{argmin}}\;\bigg(-\frac{\lambda_2}{m}{Q}(\theta, \theta_b)\bigg)\,,\; \text{if ground-truth latent states are not available,}
	\end{align}
	where the supervised loss $\mathcal{L}(\theta)$ can be selected from supervised losses introduced in previous sections and $\lambda_1$, $\lambda_2$ are hyperparameters.
	
	\subsection{Variational objectives}
	\label{subsec:vi_loss}
	In variational inference (VI) problems, using the evidence lower bound (ELBO) as a surrogate objective for the maximum likelihood estimation (MLE) enables one to perform model learning and proposal learning simultaneously and has demonstrated its effectiveness in the literature~\cite{kingma2014auto,rezende2015variational,burda2016importance,jordan1999introduction,beal2003variational,rezende2014stochastic}. There are two main components in VI, i.e., a parametric family of variational distributions and the ELBO as the training objective. Following the framework of VI, in~\cite{naesseth2018variational,le2018auto,maddison2017filtering}, the components of particle filters were optimised with a filtering variational ELBO, which was proposed based on the estimator of the marginal likelihood given by particle filters.
	
	To present the filtering variational objective, we first introduce a unified perspective on ELBOs used in VI. Given a set of observations $\{y_t\}_{t=1}^{T}$, the ELBO of the logarithmic marginal likelihood $\log p(y_{1:T};\theta)$ can be derived by:
	\begin{align}
		\label{eq:elbo_1}
		\log p(y_{1:T};\theta)&=\log \int_{\mathcal{X}^{T}} p(y_{1:T}, x_{0:T};\theta)\text{d}x_{0:T}\\
		\label{eq:elbo_2}
		&=\log \int_{\mathcal{X}^{T}} \frac{p(y_{1:T}, x_{0:T};\theta)}{q(x_{0:T}| y_{1:T};\phi)}q(x_{0:T}| y_{1:T};\phi)\text{d}x_{0:T}\\
		\label{eq:elbo_3}
		&=\log \mathbb{E}_q\big[\frac{p(y_{1:T}, x_{1:T};\theta)}{q(x_{0:T}| y_{1:T};\phi)}\big]\\
		&\geq \mathbb{E}_q\big[\log\frac{p(y_{1:T}, x_{1:T};\theta)}{q(x_{0:T}| y_{1:T};\phi)}\big]\,,
		\label{eq:elbo_4}
	\end{align}
	where Jensen's inequality is applied from Eq.~\eqref{eq:elbo_3} to Eq.~\eqref{eq:elbo_4}, and $q(x_{0:T}| y_{1:T};\phi)$ is the proposal distribution. It can be observed from Eq.~\eqref{eq:elbo_1} to Eq.~\eqref{eq:elbo_4} that as long as we can derive an unbiased estimator of the observation likelihood $p(y_{1:T};\theta)$, e.g. $\mathbb{E}_q\big[\frac{p(y_{1:T}, x_{0:T};\theta)}{q(x_{0:T}| y_{1:T};\phi)}\big]$ in Eq.~\eqref{eq:elbo_3}, we can apply Jensen's inequality to obtain an ELBO of the logarithmic marginal likelihood $\log p(y_{1:T};\theta)$. Following this pattern, the importance weighted auto-encoder proposes to use the estimator $\frac{1}{N_p}\sum_{i=1}^{N_p}\frac{p(y_{1:T}, x^i_{0:T};\theta)}{q(x^i_{0:T}| y_{1:T};\phi)}$ by drawing multiple samples from the proposal:
	\begin{align}
		\log p(y_{1:T};\theta)&
		=\log \mathbb{E}_{\mathcal{Q}_{\text{IWAE}}}\big[\frac{1}{N_p}\sum_{i=1}^{N_p}\frac{p(y_{1:T}, x^i_{0:T};\theta)}{q(x^i_{0:T}| y_{1:T};\phi)}\big]\\
		\label{eq:elbo_iwae}
		&\geq \mathbb{E}_{\mathcal{Q}_{\text{IWAE}}}\bigg[\log\big(\frac{1}{N_p}\sum_{i=1}^{N_p}\frac{p(y_{1:T}, x^i_{0:T};\theta)}{q(x^i_{0:T}| y_{1:T};\phi)}\big)\bigg]\,,
	\end{align}
	where $\mathcal{Q}_{\text{IWAE}}=\prod_{i=1}^{N_p}q(x^i_{0:T}| y_{1:T};\phi)$. It was proved in \cite{burda2016importance} that with $N_p\geq 1$, Eq.~\eqref{eq:elbo_iwae} is a tighter bound than the ELBO defined by Eq.~\eqref{eq:elbo_4}. Similarly, the filtering ELBO for SMC methods can be derived as:
	\begin{align}
		\log p(y_{1:T};\theta)&\geq\mathbb{E}_{\mathcal{Q}_\text{SMC}}\big[\log\hat{p}(y_{1:T};\theta)\big]\\
		&=\int_{\mathcal{X}_{0:T}} \mathcal{Q}_\text{SMC}(x_{0:T}^{1:N_p}, A_{1:T-1}^{1:N_p})\log\hat{p}(y_{1:T};\theta)\text{d}x^{1:N_p}\,,
	\end{align}
	where $\mathcal{Q}_\text{SMC}(x_{0:T}^{1:N_p}, A_{0:T-1}^{1:N_p})$ is the sampling distribution of SMC~\cite{le2018auto}, $x_{0:T}^{1:N_p}$ and $A_{0:T-1}^{1:N_p}$ respectively refer to samples from the proposal distribution and resampling indices, and $\hat{p}(y_{1:T};\theta)$ is the SMC estimates of the marginal likelihood:
	\begin{align}
		\hat{p}(y_{1:T};\theta)=\prod_{t=0}^{T}\bigg[\frac{1}{N_p}\sum_{i=1}^{N_p}{w}_t^i\bigg]\,.\nonumber
	\end{align}

\section{Examples of pseudocode for differentiable particle filters}
\label{sec:examples}
    With the above sections introducing different components of differentiable particle filters, we provide in this section pseudocode for two recent differentiable particle filtering frameworks to reveal sufficient implementation details for interested readers to implement differentiable particle filters in their applications.
\subsection{Particle filter networks}
    We first present the implementation detail of the particle filter network (PFNet), a differentiable particle filter proposed in~\cite{karkus2018particle}. The PFNet was developed for a robot localisation task, where the goal in the task is to localise a robot with image observations collected from onboard cameras. The soft resampler introduced in Section~\ref{subsec:soft_resampling} was applied in the PFNet. The latent state in the PFNet is defined as $x_t=[s^{(1)}_t, s^{(2)}_t, \eta_t]$, where $[s^{(1)}_t, s^{(2)}_t]$ and $\eta_t$ are the location and the orientation of the robot respectively. At each time step, the actions $a_t=[v_t^{(1)},v_t^{(2)}, \omega_t]$ of the robot are given. Given actions, the transition of the robot location is as follows:
    \begin{gather}
    \label{eq:PFNet_transition_1}
        \hat{\eta}_{t} = \eta_{t} + \alpha_t^{(3)}\,,\;\eta_{t+1} = \hat{\eta}_{t} + \omega_{t}\,,\\
        s^{(1)}_{t+1}=s^{(1)}_t+v_t^{(1)}\cos(\hat{\eta}_{t})+v_t^{(2)}\sin(\hat{\eta}_{t})+ \alpha_t^{(1)}\,,\\
        s^{(2)}_{t+1}=s^{(2)}_t+v_t^{(1)}\sin(\hat{\eta}_{t})-v_t^{(2)}\cos(\hat{\eta}_{t})+ \alpha_t^{(2)}\,,
    \label{eq:PFNet_transition_3}        
    \end{gather}
    where $\alpha_t^{(1)}\sim\mathcal{N}(0, \sigma_1^2)$, $\alpha_t^{(2)}\sim\mathcal{N}(0, \sigma_2^2)$, and $\alpha_t^{(3)}\sim\mathcal{N}(0, \sigma_1^3)$ are the noise term of the dynamic model. The PFNet uses its dynamic model to propose new particles, i.e. the PFNet is a bootstrap particle filter. Global 2D maps of different environments are given, and a spatial transformer network is used in the PFNet to transform global maps into local 2D maps according to the robot's location and orientation. The local 2D maps are then compared with the onboard 3D images from robot cameras to evaluate the conditional likelihood of observations given states, as introduced in Section~\ref{subsec:measurement_nn}. The loss function used to train the PFNet is the mean square error (MSE) between the predicted state and the ground-truth latent state. 
    
    The pseudocode for the PFNet can be found in Algorithm~\ref{PFNet}. The PFNet has achieved impressive performance in a challenging robot localisation task and generalises well to unseen environments. 
    
    \begin{algorithm}[t]
        \begin{algorithmic}[1]
            \caption{Particle filter network (PFNet)~\cite{karkus2018particle}.}\label{PFNet}
            \STATE \textbf{Inputs}: Initial distribution $\pi(x_0;\theta)$, resampling threshold $\text{ESS}_{\text{min}}$, particle number $N_p$, dynamic model $p(x_{t}| x_{t-1};\theta)$, soft resampler $\mathcal{R}_\lambda(\{W_t^i\}_{i\in[N_p]})$ that outputs indices $A_t^i$ of selected particles, spatial transformer network $\textbf{ST}_\theta(\cdot)$, global 2D map $\mathbb{M}$, local 2D map generated by the spatial transformer network $M_{x_t^i}$, scalar function $h_\theta(\cdot)$ parameterised by neural networks, ground-truth robot location $x_t^*=[s^{(1)*}_t, s^{(2)*}_t, \eta^*_t]$, hyperparameters $\lambda$ and $\xi$, learning rate $\zeta$, $[N_p]:=\{1,\cdots,N_p\}$;
            \STATE \textbf{Initialisation}: Sample $x_0^i\overset{\text{i.i.d}}{\sim}\pi(x_0;\theta)$ for $\forall i\in[N_p]$;
            \STATE Set weights $w_0^i=1$ for $\forall i\in[N_p]$;
            \STATE Normalise weights $W_0^i={w_0^i}/{\sum_{n=1}^{N_p}w_0^n}$ for $\forall i\in[N_p]$;
            \STATE Compute the effective sample size: $\text{ESS}_0(\{W_{0}^i\}_{i\in [N_p]})=\frac{1}{\sum_{i=1}^{N_p}(W_{0}^i)^2}$;
            \FOR {$t=1$ to $T$}
            \IF {$\text{ESS}_t(\{W_{t-1}^i\}_{i\in [N_p]}) < \text{ESS}_{\text{min}}$}
            \STATE $\tilde{W}_{t-1}^i=\lambda W_{t-1}^i+(1-\lambda)\frac{1}{N_p}$ (Eq.~\eqref{eq:soft_weights}).
            \STATE $A_t^{1:N_p}\leftarrow$ $\mathcal{R}(\{\tilde{W}_{t-1}^i\}_{i\in[N_p]})$, $\tilde{{w}}_{t-1}^i\leftarrow\frac{W_{t-1}^i}{\tilde{W}_{t-1}^i}$ for $\forall i \in[N_p]$ (Eq.~\eqref{eq:soft_resampling_weights_corrected});
            \ELSE
            \STATE $A_t^{i}\leftarrow i$, $\tilde{{w}}_{t-1}^i\leftarrow {{w}}_{t-1}^i$ for $\forall i \in[N_p]$;
            \ENDIF
            \STATE Sample $x_t^i\overset{\text{i.i.d}}{\sim}p(x_t|x^{A_t^i}_{t-1};\theta)$ for $\forall i\in[N_p]$ using Eqs.~\eqref{eq:PFNet_transition_1} --~\eqref{eq:PFNet_transition_3};
            \STATE Generate local 2D maps : $M_{x_t^i}=\textbf{ST}_\theta(\mathbb{M}, x_t^i)$;
		\STATE Compute unnormalised conditional likelihood: $p(y_t| x_t^i;\theta) \propto h_\theta(y_t, M_{x_t^i})$;
            \STATE Update weights $w_t^i=\tilde{w}_{t-1}^i h_\theta(y_t, M_{x_t^i})$;
            \STATE Normalise weights: $W_t^i={w_t^i}/{\sum_{n=1}^{N_p}w_t^n}$ for $\forall i\in[N_p]$;
            \STATE Estimate robot location $\bar{x}_t:=[\bar{s}^{(1)}_t, \bar{s}^{(2)}_t, \bar{\eta}_t]$: $\bar{x}_t=\sum_{i=1}^{N_p}W_t^i x_t^i$.
            \STATE Compute the effective sample size: $\text{ESS}_t(\{W_{t}^i\}_{i\in [N_p]})=\frac{1}{\sum_{i=1}^{N_p}(W_{t}^i)^2}$;
            % 		\ELSE
            % 		\STATE  $s_t^i=\tilde{s}_t^i$ for $i=1,...,N_p$;
            \ENDFOR
            \STATE Loss function $\mathcal{L}=\sum_{t=0}^T (\bar{s}^{(1)}_t-s^{(1)*}_t)^2+(\bar{s}^{(2)}_t-s^{(2)*}_t)^2 +\xi (\bar{\eta}_t-\eta^*_t)^2$;
            \STATE Update $\theta$ by gradient descent: $\theta\leftarrow\theta-\zeta\nabla_{\theta}\mathcal{L}$.
        \end{algorithmic}
    \end{algorithm}

    \subsection{Normalising flow-based differentiable particle filters}
    Differentiable particle filters proposed in~\cite{chen2021differentiable} and~\cite{chen2022conditional} are constructed using normalising flows. In~\cite{chen2021differentiable} normalising flows are used to construct dynamic models and proposal distributions, and in~\cite{chen2022conditional} normalising flows are used to construct measurement models. Measurement models built with conditional normalising flows can be trained with likelihood-based training objectives as they admit valid probability densities by construction. Additionally, the normalising flow-based differentiable particle filters presented in Algorithm~\ref{alg:nf_dpfs} also provide a flexible mechanism to build complex dynamic models and proposal distributions. The pseudocode presented in Algorithm~\ref{alg:nf_dpfs} provides the implementation details of a differentiable particle filter whose dynamic model, proposal distribution, and measurement model are built with (conditional) normalising flows. In Algorithm~\ref{alg:nf_dpfs}, the entropy-regularised OT resampler is employed, following the setup in~\cite{chen2022conditional}.
    	\begin{algorithm}[ht!]
		\begin{algorithmic}[1]
			\caption{Normalising flow-based differentiable particle filters~\cite{chen2021differentiable,chen2022conditional}.}\label{alg:nf_dpfs}
			\STATE \textbf{Inputs}: {Initial distribution $\pi(x_0;\theta)$, resampling threshold $\text{ESS}_{\text{min}}$, particle number $N_p$, dynamic model $p(x_{t}| x_{t-1};\theta)$, proposal distribution $q(x_{t}|y_t, x_{t-1};\phi)$, base dynamic model $\tilde{p}(\tilde{x}_{t}| x_{t-1};\theta)$, entropy-regularised OT resampler $\mathcal{R}_\xi(\{x_t^i\}_{i\in[N_p]},\{W_t^i\}_{i\in[N_p]})$, dynamic model normalising flow $\mathcal{T}_\theta(\cdot)$, proposal normalising flow $\mathcal{G}_\phi(\cdot)$, measurement normalising flow $\mathcal{F}_\theta(\cdot)$, neural network $E_\theta(\cdot)$, learning rate $\zeta$, $p_z(\cdot)$ probability density function of standard Gaussian distribution, hyperparameters $\xi$, $[N_p]:=\{1,\cdots,N_p\}$;}
			
            \STATE \textbf{Initialisation}: Sample $x_0^i\overset{\text{i.i.d}}{\sim}\pi(x_0;\theta)$ for $\forall i\in[N_p]$;
            \STATE Set weights $w_0^i=1$ for $\forall i\in[N_p]$;
            \STATE Normalise weights $W_0^i={w_0^i}/{\sum_{n=1}^{N_p}w_0^n}$ for $\forall i\in[N_p]$;
            \STATE Compute the effective sample size: $\text{ESS}_0(\{W_{0}^i\}_{i\in [N_p]})=\frac{1}{\sum_{i=1}^{N_p}(W_{0}^i)^2}$;
			\FOR {$t=1$ to $T$}
                \IF {$\text{ESS}_t(\{W_{t-1}^i\}_{i\in [N_p]}) <\text{ESS}_{\text{min}}$}
			\STATE $\{{x}_{t-1}^{i}\}_{i=1}^{N}\leftarrow$ $\mathcal{R}_\xi(\{x_{t-1}^i\}_{i\in[N_p]},\{W_{t-1}^i\}_{i\in[N_p]})$, $\tilde{w}_{t-1}^i\leftarrow {1}$, $\forall i \in[N_p]$;
			\ELSE
			\STATE $\tilde{w}_{t-1}^i\leftarrow {w}_{t-1}^i$, $\forall i \in[N_p]$;
			% \STATE ${{w}}_{t-1}^i\leftarrow {{w}}_{t-1}^i$, $\forall i \in[N_p]$;
			\ENDIF
			\STATE Sample from the dynamic model (Eq.~\eqref{eq:dynamic_model}): \\$\check{x}_t^i=\mathcal{T}_\theta(\tilde{x}_t^i)\overset{\text{i.i.d}}{\sim} p(\check{x}_t | x_{t-1}^i;\theta)$ where $\tilde{x}_t^i\overset{\text{i.i.d}}{\sim} \tilde{p}(\tilde{x}_t | x_{t-1}^i;\theta)$, $\forall i\in[N_p]$;
			%			\STATE Apply the dynamic normalising flow: $\check{x}_t^i=\mathcal{T}_\theta(\breve{x}_t^i)\sim p(\check{x}_t|x_{t-1};\theta)$, $\forall i\in[N_p]$;
			\STATE Generate particles (Eq.~\eqref{eq:proposal_cnf}):
			$x_t^i=\mathcal{G}_\phi(\check{x}_t^i, y_t)\sim q(x_{t}| y_{t},x_{t-1}^i;\phi)$, $\forall i\in[N_p]$;
            \STATE Compute particle densities in the proposal distribution (Eq.~\eqref{eq:cnf_proposal_density}): \\$q(x_{t}^i| y_{t}, x_{t-1}^i;\phi)=\frac{p\left(\check{x}_{t}^{i} | x_{t-1}^{i}; \theta\right)}{|\operatorname{det}J_{\mathcal{G}_{\phi}}\left(\check{x}_{t}^{i}\right)|}=\frac{\tilde{p}\left(\tilde{x}_{t}^{i} | x_{t-1}^{i}; \theta\right)}{|\operatorname{det}J_{\mathcal{G}_{\phi}}\left(\check{x}_{t}^{i}\right)||\operatorname{det}J_{\mathcal{T}_{\phi}}(\tilde{x}_{t}^{i})|}$, $\forall i\in[N_p]$;
            \STATE Compute particle densities in the dynamic model  (Eq.~\eqref{eq:density_nf}):\\ $p(x_t^i | x_{t-1}^i;\theta)=\frac{\tilde{p}(\mathcal{T}^{-1}({x}_t^i) | x_{t-1}^i;\theta)}{|\operatorname{det}J_{\mathcal{T_\theta}}(\mathcal{T}^{-1}({x}_t^i) )|}$, $\forall i\in[N_p]$;
			\STATE Compute conditional likelihood of observations given states (Eq.~\eqref{eq:cnf_measurement}):\\$p(y_t| x_t^i;\theta)=p_z(z_t^i)\bigg|\operatorname{det}\frac{\partial z_t^i}{\partial y_t}\bigg|$, with $z_t^i=\mathcal{F}_\theta(y_t, x_t^i)$, $\forall i\in[N_p]$;
			\STATE Update weights:
            $w_t^i=\tilde{w}_{t-1}^i \frac{p(y_{t}| x_{t}^{i}; \theta) p(x_t^i| {x}_{t-1}^i;\theta)}{q(x_t^i| y_{t},{x}_{t-1}^i;\phi)}$, $\forall i\in[N_p]$;
                \STATE Normalise weights: $W_t^i={w_t^i}/{\sum_{n=1}^{N_p}w_t^n}$ for $\forall i\in[N_p]$;
                \STATE Compute the effective sample size: $\text{ESS}_t(\{W_{t}^i\}_{i\in [N_p]})=\frac{1}{\sum_{i=1}^{N_p}(W_{t}^i)^2}$;
			\ENDFOR
			\STATE Compute the overall loss function $\mathcal{L}(\theta, \phi)$;
			\STATE Update $\theta$ and $\phi$:
			$\theta\leftarrow \theta-\xi\nabla_\theta\mathcal{L}(\theta, \phi)\,, \phi\leftarrow \phi-\xi\nabla_\phi\mathcal{L}(\phi, \phi)$.
			
		\end{algorithmic}
	\end{algorithm}
 
	\section{Discussion \& Summary}
	\label{sec:conclusion}
	In this paper, we reviewed recent advances in differentiable particle filters, with a particular focus on five core components of differentiable particle filters: dynamic models, measurement models, proposal distributions, differentiable resampling techniques, and training objectives. 
	Specifically, the above modules of differentiable particle filters can be summarised as follows:
	\begin{enumerate}
		\item Dynamic models used in many differentiable particle filter frameworks are Gaussian, where the transition of states is simulated by a Gaussian distribution whose mean and variance are predicted by a function of previous states. Normalising flows can be used to transform simple distributions, e.g. Gaussian, into arbitrarily complex distributions under mild conditions, and they also enable the evaluation of densities of transformed particles. Dynamic models in several works are built with observations, and have shown competitive performances in several examined scenarios. However, this class of dynamic models does not fit into the problem setup of state-space models we consider in Bayesian filtering problems.
		\item For the proposal distribution of differentiable particle filters, we can employ conditional normalising flows to incorporate information from observations into the construction of proposal distributions, with a straightforward evaluation of the proposal density.
		\item In cases where the relationship between states and observations can be built with known distribution families, measurement models are often differentiable by themselves. In more complex scenarios, differentiable particle filters resort to build connections between observations and states in a feature space. Typical examples include: (i) measurement models as scalar functions built with neural networks, where the conditional likelihood of observation is given by neural networks with observation and state features as inputs; (ii) measurement models as feature similarities, where the conditional likelihood of observation is evaluated as the similarity between the observation feature and the state feature; (iii) measurement models built with conditional normalising flows, where observations are transformed into Gaussian samples through invertible neural networks and the conditional likelihood of observation is computed through the change of variable formula.
		\item Different techniques for differentiable resampling have been proposed to address differentiability of the resampling step. In the soft resampling method,  the gradients of the importance weights before resampling w.r.t. previous weights are non-zero. Soft resampling is not truly differentiable as the non-differentiable part of resampling is ignored. A fully-differentiable resampling scheme is achieved by considering the resampling operation through solving entropy-regularised optimal transport problems. The regularisation term in the OT-resampler introduces biases into resampling outcomes, and the biases only vanish if the regularisation coefficient approaches zero. A transformer-based resampler, particle transformer, was proposed based on the recent development of the transformer architecture and self-attention mechanism. One limitation of the particle transformer is that a particle transformer trained for one specific task may not be suitable for other tasks.
		\item The training objective of differentiable particle filters include supervised losses, semi-supervised losses, and variational objectives. Supervised losses require ground-truth latent states to train differentiable particle filters. One type of semi-supervised losses includes a pseudo-likelihood loss to optimise parameters of differentiable particle filters when ground-truth latent states are unavailable. Variational objectives adopt an evidence lower bound (ELBO) of the marginal log-likelihood of observations as a surrogate objective for training differentiable particle filters.
	\end{enumerate}
        Compared with traditional particle filters, differentiable particle filters are particularly suitable for performing sequential Bayesian inference in complex environments where neither the structure nor the parameters of the system of interest are known. In such cases, traditional particle filters require hand-crafted designs of the components of particle filters, which can be challenging, often rely on extensive domain knowledge, and can lead to significant delays in the deployment of the filtering algorithm. The trade-off is that differentiable particle filters have to be trained before they can produce reasonable filtering results and therefore are more computationally expensive to train than traditional approaches. Nonetheless, differentiable particle filters provide a data-adaptive framework that holds the promise to significantly accelerate the development and adoption of particle filters in real-world sequential Bayesian filtering applications, with many open research questions awaiting to be explored.
	\bibliographystyle{IEEEtran}
	\bibliography{ref_literature_review.bib}

\begin{thebibliography}{100}
\providecommand{\url}[1]{#1}
\csname url@samestyle\endcsname
\providecommand{\newblock}{\relax}
\providecommand{\bibinfo}[2]{#2}
\providecommand{\BIBentrySTDinterwordspacing}{\spaceskip=0pt\relax}
\providecommand{\BIBentryALTinterwordstretchfactor}{4}
\providecommand{\BIBentryALTinterwordspacing}{\spaceskip=\fontdimen2\font plus
\BIBentryALTinterwordstretchfactor\fontdimen3\font minus
  \fontdimen4\font\relax}
\providecommand{\BIBforeignlanguage}[2]{{%
\expandafter\ifx\csname l@#1\endcsname\relax
\typeout{** WARNING: IEEEtran.bst: No hyphenation pattern has been}%
\typeout{** loaded for the language `#1'. Using the pattern for}%
\typeout{** the default language instead.}%
\else
\language=\csname l@#1\endcsname
\fi
#2}}
\providecommand{\BIBdecl}{\relax}
\BIBdecl

\bibitem{kalman1960new}
R.~E. Kalman, ``A new approach to linear filtering and prediction problems,''
  \emph{J .Basic Eng.}, vol.~82, no.~1, pp. 33--45, 1960.

\bibitem{anderson1979optimal}
B.~Anderson and J.~B. Moore, ``Optimal filtering,'' \emph{Prentice-Hall Inform.
  and Syst. Sci. Ser.}, 1979.

\bibitem{daum2015extended}
F.~E. Daum, ``Extended {K}alman filters.'' \emph{Encyclo. Syst. and Control},
  2015.

\bibitem{wan2000unscented}
E.~A. Wan and R.~Van Der~Merwe, ``The unscented kalman filter for nonlinear
  estimation,'' in \emph{Proc. IEEE Adapt. Syst. Signal. Process. Commun.
  Control Sympos.}, Oct., Lake Louise, Canada 2000.

\bibitem{bucy1971digital}
R.~S. Bucy and K.~D. Senne, ``Digital synthesis of non-linear filters,''
  \emph{Automatica}, vol.~7, no.~3, pp. 287--298, 1971.

\bibitem{doucet2001introduction}
A.~Doucet, D.~N. Freitas, and N.~Gordon, ``An introduction to sequential
  {M}onte {C}arlo methods,'' in \emph{Sequential Monte Carlo methods in
  practice}.\hskip 1em plus 0.5em minus 0.4em\relax Springer, 2001, pp. 3--14.

\bibitem{gordon1993novel}
N.~Gordon, D.~Salmond, and A.~Smith, ``{Novel approach to
  nonlinear/non-{Gaussian} {Bayesian} state estimation},'' in \emph{IEE Proc. F
  (Radar and Signal Process.)}, vol. 140, 1993, pp. 107--113.

\bibitem{djuric2003particle}
P.~M. Djuri\'c, J.~H. Kotecha, J.~Zhang, Y.~Huang, T.~Ghirmai, M.~F. Bugallo,
  and J.~Miguez, ``Particle filtering,'' \emph{IEEE Signal Process. Mag.},
  vol.~20, no.~5, pp. 19--38, 2003.

\bibitem{doucet2001sequential}
A.~Doucet, N.~De~Freitas, N.~J. Gordon \emph{et~al.}, \emph{{Sequential Monte
  Carlo} methods in practice}.\hskip 1em plus 0.5em minus 0.4em\relax Springer,
  2001, vol.~1, no.~2.

\bibitem{del2000branching}
P.~Del~Moral and L.~Miclo, ``Branching and interacting particle systems.
  approximations of {Feynman-Kac} formulae with applications to non-linear
  filtering,'' \emph{S{\'e}min. Probab. Strasbourg}, vol.~34, pp. 1--145, 2000.

\bibitem{crisan2002survey}
D.~Crisan and A.~Doucet, ``A survey of convergence results on particle
  filtering methods for practitioners,'' \emph{IEEE Trans. Signal Process.},
  vol.~50, no.~3, pp. 736--746, 2002.

\bibitem{del1998measure}
P.~Del~Moral, ``Measure-valued processes and interacting particle systems.
  application to nonlinear filtering problems,'' \emph{Ann. Appl. Probab.},
  vol.~8, no.~2, pp. 438--495, 1998.

\bibitem{elvira2017adapting}
V.~Elvira, J.~M{\'\i}guez, and P.~M. Djuri{\'c}, ``Adapting the number of
  particles in sequential {Monte Carlo} methods through an online scheme for
  convergence assessment,'' \emph{IEEE Trans. Signal Process.}, vol.~65, no.~7,
  pp. 1781--1794, 2017.

\bibitem{elvira2021performance}
V.~Elvira, J.~Miguez, and P.~M. Djuri{\'c}, ``On the performance of particle
  filters with adaptive number of particles,'' \emph{Stat. Comput.}, vol.~31,
  pp. 1--18, 2021.

\bibitem{pitt1999filtering}
M.~K. Pitt and N.~Shephard, ``Filtering via simulation: Auxiliary particle
  filters,'' \emph{J. Amer. Statist. Assoc.}, vol.~94, no. 446, pp. 590--599,
  1999.

\bibitem{elvira2019elucidating}
V.~Elvira, L.~Martino, M.~F. Bugallo, and P.~M. Djuri\'c, ``Elucidating the
  auxiliary particle filter via multiple importance sampling,'' \emph{IEEE
  Signal Process. Mag.}, vol.~36, no.~6, pp. 145--152, 2019.

\bibitem{branchini2021optimized}
N.~Branchini and V.~Elvira, ``Optimized auxiliary particle filters: adapting
  mixture proposals via convex optimization,'' in \emph{Proc. Conf. Uncertain.
  Artif. Intell. (UAI)}, 2021, pp. 1289--1299.

\bibitem{kotecha2003gaussian}
J.~H. Kotecha and P.~M. Djuri\'c, ``Gaussian sum particle filtering,''
  \emph{IEEE Trans. Signal Process.}, vol.~51, no.~10, pp. 2602--2612, 2003.

\bibitem{kotecha2001gaussian}
------, ``Gaussian sum particle filtering for dynamic state-space models,'' in
  \emph{Proc. IEEE Int. Conf. Acoust. Speech Signal Process.}, Salt Lake City,
  USA, May 2001.

\bibitem{kotecha2003gaussian_particle}
------, ``Gaussian particle filtering,'' \emph{IEEE Trans. Signal Process.},
  vol.~51, no.~10, pp. 2592--2601, 2003.

\bibitem{van2000unscented}
R.~Van Der~Merwe, A.~Doucet, N.~De~Freitas, and E.~Wan, ``The unscented
  particle filter,'' \emph{Proc. Adv. Neur. Inf. Process. Sys. (NeurIPS)}, Dec.
  2000.

\bibitem{rui2001better}
Y.~Rui and Y.~Chen, ``Better proposal distributions: Object tracking using
  unscented particle filter,'' in \emph{IEEE Conf. Comp. Vis. and Pat. Recog.
  (CVPR)}.\hskip 1em plus 0.5em minus 0.4em\relax Kauai, USA.: IEEE, Dec. 2001.

\bibitem{julier2004unscented}
S.~J. Julier and J.~K. Uhlmann, ``Unscented filtering and nonlinear
  estimation,'' \emph{Proc. IEEE}, vol.~92, no.~3, pp. 401--422, 2004.

\bibitem{doucet2000rao}
A.~Doucet, N.~de~Freitas, K.~Murphy, and S.~Russell, ``{Rao-Blackwellised}
  particle filtering for dynamic {Bayesian} networks,'' in \emph{Proc. Conf.
  Uncertain. Artif. Intell. (UAI)}, Stanford, USA, 2000, pp. 176--183.

\bibitem{de2002rao}
N.~De~Freitas, ``{Rao-Blackwellised} particle filtering for fault diagnosis,''
  in \emph{Proc. IEEE Aerosp. Conf.}, vol.~4, 2002, pp. 4--4.

\bibitem{djuric2007}
P.~M. Djuri\'c, T.~Lu, and M.~F. Bugallo, ``Multiple particle filtering,'' in
  \emph{Proc. IEEE Int. Conf. Acoust. Speech Signal Process. (ICASSP)},
  Honolulu, USA, Apr. 2007.

\bibitem{djuric2013}
P.~M. Djuri\'c and M.~F. Bugallo, ``Particle filtering for high-dimensional
  systems,'' in \emph{Proc. Comput. Adv. Multi-Sensor Adapt. Process.
  (CAMSAP)}, Saint Martin, France, Dec. 2013.

\bibitem{bunch2016approximations}
P.~Bunch and S.~Godsill, ``Approximations of the optimal importance density
  using gaussian particle flow importance sampling,'' \emph{J. Amer. Statist.
  Assoc.}, vol. 111, no. 514, pp. 748--762, 2016.

\bibitem{li2017particle}
Y.~Li and M.~Coates, ``Particle filtering with invertible particle flow,''
  \emph{IEEE Trans. Signal Process.}, vol.~65, no.~15, pp. 4102--4116, 2017.

\bibitem{heng2021gibbs}
J.~Heng, A.~Doucet, and Y.~Pokern, ``Gibbs flow for approximate transport with
  applications to {Bayesian} computation,'' \emph{J. R. Stat. Soc. Ser. B.
  Stat. Methodol}, vol.~83, no.~1, pp. 156--187, 2021.

\bibitem{zhang2017multi}
T.~Zhang, C.~Xu, and M.-H. Yang, ``Multi-task correlation particle filter for
  robust object tracking,'' in \emph{Proc. IEEE Conf. Comput. Vis. and Pattern
  Recogn. (CVPR)}, Honolulu, Hawaii, July 2017.

\bibitem{malik2011particle}
S.~Malik and M.~K. Pitt, ``Particle filters for continuous likelihood
  evaluation and maximisation,'' \emph{J. Econometrics}, vol. 165, no.~2, pp.
  190--209, 2011.

\bibitem{gunatilake2022novel}
A.~Gunatilake, S.~Kodagoda, and K.~Thiyagarajan, ``A novel {UHF-RFID} dual
  antenna signals combined with {G}aussian process and particle filter for
  in-pipe robot localization,'' \emph{IEEE Robot. Autom. Lett.}, vol.~7, no.~3,
  pp. 6005--6011, 2022.

\bibitem{van2019particle}
P.~J. Van~Leeuwen, H.~R. K{\"u}nsch, L.~Nerger, R.~Potthast, and S.~Reich,
  ``Particle filters for high-dimensional geoscience applications: A review,''
  \emph{Q. J. R. Meteorol. Soc.}, vol. 145, no. 723, pp. 2335--2365, 2019.

\bibitem{pozna2022hybrid}
C.~Pozna, R.-E. Precup, E.~Horv{\'a}th, and E.~M. Petriu, ``Hybrid particle
  filter--particle swarm optimization algorithm and application to fuzzy
  controlled servo systems,'' \emph{IEEE Trans. Fuzzy Syst.}, vol.~30, no.~10,
  pp. 4286--4297, 2022.

\bibitem{kantas2009overview}
N.~Kantas, A.~Doucet, S.~S. Singh, and J.~M. Maciejowski, ``An overview of
  sequential {M}onte {C}arlo methods for parameter estimation in general
  state-space models,'' \emph{IFAC Proc. Vol.}, vol.~42, no.~10, pp. 774--785,
  2009.

\bibitem{kantas2015particle}
N.~Kantas, A.~Doucet, S.~S. Singh, J.~Maciejowski, and N.~Chopin, ``On particle
  methods for parameter estimation in state-space models,'' \emph{Stat. Sci.},
  vol.~30, no.~3, pp. 328--351, 2015.

\bibitem{andrieu2010particle}
C.~Andrieu, A.~Doucet, and R.~Holenstein, ``Particle {Markov chain Monte Carlo}
  methods,'' \emph{J. R. Stat. Soc. Ser. B. Stat. Methodol.}, vol.~72, no.~3,
  pp. 269--342, 2010.

\bibitem{lindsten2014particle}
F.~Lindsten, M.~I. Jordan, and T.~B. Schon, ``Particle gibbs with ancestor
  sampling,'' \emph{J. Mach. Learn. Res.}, vol.~15, pp. 2145--2184, 2014.

\bibitem{chopin2013smc2}
N.~Chopin, P.~E. Jacob, and O.~Papaspiliopoulos, ``{SMC2}: an efficient
  algorithm for sequential analysis of state space models,'' \emph{J. R. Stat.
  Soc. Ser. B. Stat. Methodol.}, vol.~75, no.~3, pp. 397--426, 2013.

\bibitem{perez2018probabilistic}
S.~P{\'e}rez-Vieites, I.~P. Mari{\~n}o, and J.~M{\'\i}guez, ``Probabilistic
  scheme for joint parameter estimation and state prediction in complex
  dynamical systems,'' \emph{Phys. Rev. E}, vol.~98, no.~6, p. 063305, 2018.

\bibitem{crisan2018nested}
D.~Crisan and J.~M{\'I}guez, ``Nested particle filters for online parameter
  estimation in discrete-time state-space {Markov} models,'' \emph{Bernoulli},
  vol.~24, no.~4A, pp. 3039--3086, 2018.

\bibitem{perez2021nested}
S.~P{\'e}rez-Vieites and J.~M{\'\i}guez, ``Nested {Gaussian} filters for
  recursive {Bayesian} inference and nonlinear tracking in state space
  models,'' \emph{Signal Proces.}, vol. 189, p. 108295, 2021.

\bibitem{chouzenoux2020graphem}
E.~Chouzenoux and V.~Elvira, ``{GraphEM: EM algorithm for blind Kalman
  filtering under graphical sparsity constraints},'' in \emph{Proc. IEEE Int.
  Conf. Acoust. Speech Signal Process. (ICASSP)}, May 2020.

\bibitem{elvira2022graphical}
V.~Elvira and {\'E}.~Chouzenoux, ``Graphical inference in linear-{Gaussian}
  state-space models,'' \emph{IEEE Trans. Signal Process.}, vol.~70, pp.
  4757--4771, 2022.

\bibitem{chouzenoux2023sparse}
E.~Chouzenoux and V.~Elvira, ``Sparse graphical linear dynamical systems,''
  \emph{arXiv preprint arXiv:2307.03210}, 2023.

\bibitem{jonschkowski18}
R.~Jonschkowski, D.~Rastogi, and O.~Brock, ``Differentiable particle filters:
  end-to-end learning with algorithmic priors,'' in \emph{Proc. Robot.: Sci.
  and Syst. (RSS)}, Pittsburgh, Pennsylvania, July 2018.

\bibitem{karkus2018particle}
P.~Karkus, D.~Hsu, and W.~S. Lee, ``Particle filter networks with application
  to visual localization,'' in \emph{Proc. Conf. Robot Learn. (CoRL)}, Zurich,
  Switzerland, Oct 2018.

\bibitem{wen2021end}
H.~Wen, X.~Chen, G.~Papagiannis, C.~Hu, and Y.~Li, ``End-to-end semi-supervised
  learning for differentiable particle filters,'' in \emph{Proc. IEEE Int.
  Conf. Robot. Automat., (ICRA)}, Xi'an, China, May 2021.

\bibitem{chen2021differentiable}
X.~Chen, H.~Wen, and Y.~Li, ``Differentiable particle filters through
  conditional normalizing flow,'' in \emph{Proc. IEEE Int. Conf. Inf. Fusion
  (FUSION)}, Sun City, South Africa, Nov. 2021.

\bibitem{chen2022conditional}
X.~Chen and Y.~Li, ``Conditional measurement density estimation in sequential
  {Monte Carlo} via normalizing flow,'' in \emph{Proc. Euro. Sig. Process.
  Conf. (EUSIPCO)}, Belgrade, Serbia, Aug. 2022.

\bibitem{corenflos2021differentiable}
A.~Corenflos, J.~Thornton, G.~Deligiannidis, and A.~Doucet, ``Differentiable
  particle filtering via entropy-regularized optimal transport,'' in
  \emph{Proc. Int. Conf. Mach. Learn. (ICML)}, July 2021.

\bibitem{zhu2020towards}
M.~Zhu, K.~Murphy, and R.~Jonschkowski, ``Towards differentiable resampling,''
  \emph{arXiv preprint arXiv:2004.11938}, 2020.

\bibitem{elvira2021advances}
V.~Elvira and L.~Martino, ``Advances in importance sampling,'' \emph{Wiley
  StatsRef-Statistics Reference Online}, pp. 1--14, 2021.

\bibitem{hesterberg1995weighted}
T.~Hesterberg, ``Weighted average importance sampling and defensive mixture
  distributions,'' \emph{Technometrics}, vol.~37, no.~2, pp. 185--194, 1995.

\bibitem{tokdar2010importance}
S.~T. Tokdar and R.~E. Kass, ``Importance sampling: a review,'' \emph{Wiley
  Interdiscip. Rev. Comput. Stat.}, vol.~2, no.~1, pp. 54--60, 2010.

\bibitem{agapiou2017importance}
S.~Agapiou, O.~Papaspiliopoulos, D.~Sanz-Alonso, and A.~M. Stuart, ``Importance
  sampling: Intrinsic dimension and computational cost,'' \emph{Statist. Sci.},
  pp. 405--431, 2017.

\bibitem{bickel2008sharp}
P.~Bickel, B.~Li, and T.~Bengtsson, ``Sharp failure rates for the bootstrap
  particle filter in high dimensions,'' \emph{Pushing the limits of
  contemporary statistics: Contributions in honor of Jayanta K. Ghosh}, 2008.

\bibitem{chopin2020introduction}
N.~Chopin and O.~Papaspiliopoulos, \emph{An introduction to sequential {Monte
  Carlo}}.\hskip 1em plus 0.5em minus 0.4em\relax Springer, 2020.

\bibitem{li2015resampling_b}
T.~Li, M.~Bolic, and P.~M. Djuric, ``Resampling methods for particle filtering:
  classification, implementation, and strategies,'' \emph{IEEE Signal Process.
  Mag.}, vol.~32, no.~3, pp. 70--86, 2015.

\bibitem{gerber2019negative}
M.~Gerber, N.~Chopin, and N.~Whiteley, ``Negative association, ordering and
  convergence of resampling methods,'' \emph{Ann. Statist.}, vol.~47, no.~4,
  pp. 2236--2260, 2019.

\bibitem{li2015resampling}
T.-c. Li, G.~Villarrubia, S.-d. Sun, J.~M. Corchado, and J.~Bajo, ``Resampling
  methods for particle filtering: identical distribution, a new method, and
  comparable study,'' \emph{Frontiers Inform. Tech. \& Electron. Eng.},
  vol.~16, no.~11, pp. 969--984, 2015.

\bibitem{douc2005comparison}
R.~Douc and O.~Capp{\'e}, ``Comparison of resampling schemes for particle
  filtering,'' in \emph{Proc. Int. Symp. Image and Signal Process. and Anal.},
  Zagreb, Croatia, 2005.

\bibitem{bolic2004resampling}
M.~Boli{\'c}, P.~M. Djuri{\'c}, and S.~Hong, ``Resampling algorithms for
  particle filters: A computational complexity perspective,'' \emph{EURASIP J.
  Adv. Signal Process.}, vol. 2004, no.~15, pp. 1--11, 2004.

\bibitem{doucet2000sequential}
A.~Doucet, S.~Godsill, and C.~Andrieu, ``On sequential {Monte Carlo} sampling
  methods for bayesian filtering,'' \emph{Stat. Comput.}, vol.~10, no.~3, pp.
  197--208, 2000.

\bibitem{elvira2022rethinking}
V.~Elvira, L.~Martino, and C.~P. Robert, ``Rethinking the effective sample
  size,'' \emph{Int. Stat. Rev.}, vol.~90, no.~3, pp. 525--550, 2022.

\bibitem{martino2017effective}
L.~Martino, V.~Elvira, and F.~Louzada, ``Effective sample size for importance
  sampling based on discrepancy measures,'' \emph{Signal Process.}, vol. 131,
  pp. 386--401, 2017.

\bibitem{huggins2015convergence}
J.~H. Huggins and D.~M. Roy, ``Convergence of sequential {Monte Carlo} based
  sampling methods,'' \emph{arXiv preprint arXiv:1503.00966}, 2015.

\bibitem{li2020flexible}
Z.~Li, P.~Fan, and Y.~Dong, ``Flexible effective sample size based on the
  message importance measure,'' \emph{IEEE Open J. Signal Process.}, vol.~1,
  pp. 216--229, 2020.

\bibitem{doucet2003parameter}
A.~Doucet and V.~B. Tadi{\'c}, ``Parameter estimation in general state-space
  models using particle methods,'' \emph{Ann. Inst. Stat. Math.}, vol.~55,
  no.~2, pp. 409--422, 2003.

\bibitem{andrieu2005line}
C.~Andrieu, A.~Doucet, and V.~B. Tadic, ``On-line parameter estimation in
  general state-space models,'' in \emph{Proc. {IEEE} Conf. Dec. and Contr.
  (CDC)}, Seville, Spain, Dec. 2005.

\bibitem{kloss2021train}
A.~Kloss, G.~Martius, and J.~Bohg, ``How to train your differentiable filter,''
  \emph{Auto. Robot.}, vol.~45, no.~4, pp. 561--578, 2021.

\bibitem{rosato2022efficient}
C.~Rosato, L.~Devlin, V.~Beraud, P.~Horridge, T.~B. Sch{\"o}n, and S.~Maskell,
  ``Efficient learning of the parameters of non-linear models using
  differentiable resampling in particle filters,'' \emph{IEEE Trans. Signal
  Process.}, vol.~70, pp. 3676--3692, 2022.

\bibitem{williams1992simple}
R.~J. Williams, ``Simple statistical gradient-following algorithms for
  connectionist reinforcement learning,'' \emph{Mach. Learn.}, vol.~8, no. 3-4,
  pp. 229--256, 1992.

\bibitem{kingma2014auto}
D.~P. Kingma and M.~Welling, ``Auto-encoding variational {B}ayes,'' in
  \emph{Proc. Int. Conf. Learn. Represent. (ICLR)}, Scottsdale, Arizona, May
  2013.

\bibitem{reich2013nonparametric}
S.~Reich, ``A nonparametric ensemble transform method for {B}ayesian
  inference,'' \emph{SIAM J. Sci. Comput.}, vol.~35, no.~4, pp. A2013--A2024,
  2013.

\bibitem{cuturi2013sinkhorn}
M.~Cuturi, ``Sinkhorn distances: Lightspeed computation of optimal transport,''
  in \emph{Proc. Adv. Neur. Inf. Process. Sys. (NeurIPS)}, Lake Tahoe, USA,
  Dec. 2013.

\bibitem{feydy2019interpolating}
J.~Feydy \emph{et~al.}, ``Interpolating between optimal transport and mmd using
  {S}inkhorn divergences,'' in \emph{Proc. Int. Conf. Artif. Intell. Stat.
  (AISTAS)}, Naha, Japan, Apr. 2019.

\bibitem{peyre2019computational}
G.~Peyr{\'e} and M.~Cuturi, ``Computational optimal transport,''
  \emph{Foundations and Trends{\textregistered} in Machine Learning}, vol.~11,
  no. 5-6, pp. 355--607, 2019.

\bibitem{lee2019set}
J.~Lee \emph{et~al.}, ``Set transformer: A framework for attention-based
  permutation-invariant neural networks,'' in \emph{Proc. Int. Conf. Mach.
  Learn. (ICML)}, Baltimore, USA, June 2019.

\bibitem{vaswani2017attention}
A.~Vaswani \emph{et~al.}, ``Attention is all you need,'' in \emph{Proc. Adv.
  Neur. Inf. Process. Sys. (NeurIPS)}, Long Beach, USA, Dec. 2017.

\bibitem{ma2020particle}
X.~Ma, P.~Karkus, D.~Hsu, and W.~S. Lee, ``Particle filter recurrent neural
  networks,'' in \emph{Proc. AAAI Conf. Artif. Intell. (AAAI)}, New York, USA,
  Feb. 2020.

\bibitem{ma2019discriminative}
X.~Ma \emph{et~al.}, ``Discriminative particle filter reinforcement learning
  for complex partial observations,'' in \emph{Proc. Int. Conf. Learn. Rep.
  (ICLR)}, New Orleans, USA, May 2019.

\bibitem{chen2021deep}
R.~Chen, H.~Yin, Y.~Jiao, G.~Dissanayake, Y.~Wang, and R.~Xiong, ``Deep
  samplable observation model for global localization and kidnapping,''
  \emph{IEEE Robot. Autom. Lett. (RAL)}, vol.~6, no.~2, pp. 2296--2303, 2021.

\bibitem{karkus2021differentiable}
P.~Karkus, S.~Cai, and D.~Hsu, ``Differentiable slam-net: Learning particle
  slam for visual navigation,'' in \emph{Proc. IEEE Conf. Comp. Vis. and Pat.
  Recog. (CVPR)}, June 2021.

\bibitem{karkus2019differentiable}
P.~Karkus \emph{et~al.}, ``Differentiable algorithm networks for composable
  robot learning,'' in \emph{Proc. Robot.: Sci. and Syst. (RSS)}, Messe
  Freiburg, Germany, June 2019.

\bibitem{lee2020multimodal}
M.~A. Lee, B.~Yi, R.~Mart{\'\i}n-Mart{\'\i}n, S.~Savarese, and J.~Bohg,
  ``Multimodal sensor fusion with differentiable filters,'' in \emph{Proc.
  IEEE/RSJ Int. Conf. Intel. Robot. Sys. (IROS)}, Las Vegas, USA, Oct. 2020.

\bibitem{naesseth2018variational}
C.~Naesseth, S.~Linderman, R.~Ranganath, and D.~Blei, ``Variational sequential
  {Monte Carlo},'' in \emph{Proc. Int. Conf. Artif. Intel. and Stat.
  (AISTATS)}, Playa Blanca, Spain, Apr. 2018.

\bibitem{le2018auto}
T.~A. Le, M.~Igl, T.~Rainforth, T.~Jin, and F.~Wood, ``Auto-encoding sequential
  {Monte Carlo},'' in \emph{Proc. Int. Conf. Learn. Rep. (ICLR)}, Vancouver,
  Canada, Apr. 2018.

\bibitem{maddison2017filtering}
C.~J. Maddison \emph{et~al.}, ``Filtering variational objectives,'' in
  \emph{Proc. Adv. Neur. Inf. Process. Sys. (NeurIPS)}, Long Beach, USA, Dec.
  2017.

\bibitem{dupty2021pf}
M.~H. Dupty, Y.~Dong, and W.~S. Lee, ``{PF-GNN}: Differentiable particle
  filtering based approximation of universal graph representations,'' in
  \emph{Proc. Int. Conf. Learn. Represent. (ICLR)}, May 2021.

\bibitem{papamakarios2021normalizing}
G.~Papamakarios, E.~Nalisnick, D.~J. Rezende, S.~Mohamed, and
  B.~Lakshminarayanan, ``Normalizing flows for probabilistic modeling and
  inference,'' \emph{J. Mach. Learn. Res.}, vol.~22, pp. 1--64, 2021.

\bibitem{hochreiter1997long}
S.~Hochreiter and J.~Schmidhuber, ``Long short-term memory,'' \emph{Neural
  Comput.}, vol.~9, no.~8, pp. 1735--1780, 1997.

\bibitem{cho2014properties}
K.~Cho, B.~van Merri{\"e}nboer, D.~Bahdanau, and Y.~Bengio, ``On the properties
  of neural machine translation: Encoder--decoder approaches,'' \emph{Syntax,
  Semant. and Struct. in Stat. Transl.}, p. 103, 2014.

\bibitem{gama2022unrolling}
F.~Gama, N.~Zilberstein, R.~G. Baraniuk, and S.~Segarra, ``Unrolling particles:
  Unsupervised learning of sampling distributions,'' in \emph{Proc. IEEE Int.
  Conf. Acoust. Speech Signal Process. (ICASSP)}, Singapore, May 2022, pp.
  5498--5502.

\bibitem{gama2023unsupervised}
F.~Gama, N.~Zilberstein, M.~Sevilla, R.~Baraniuk, and S.~Segarra,
  ``Unsupervised learning of sampling distributions for particle filters,''
  \emph{arXiv preprint arXiv:2302.01174}, 2023.

\bibitem{winkler2019learning}
C.~Winkler, D.~Worrall, E.~Hoogeboom, and M.~Welling, ``Learning likelihoods
  with conditional normalizing flows,'' \emph{arXiv preprint arXiv:1912.00042},
  2019.

\bibitem{jaderberg2015spatial}
M.~Jaderberg, K.~Simonyan, A.~Zisserman, and K.~Kavukcuoglu, ``Spatial
  transformer networks,'' in \emph{Proc. Adv. Neur. Inf. Process. Sys.
  (NeurIPS)}, Montreal, Canada, Dec. 2015.

\bibitem{villani2003topics}
C.~Villani, \emph{Topics in optimal transportation}.\hskip 1em plus 0.5em minus
  0.4em\relax American Mathematical Society, 2003.

\bibitem{villani2008optimal}
------, \emph{Optimal Transport: old and new}.\hskip 1em plus 0.5em minus
  0.4em\relax Berlin, Germany: Springer Science \& Business Media, 2008, vol.
  338.

\bibitem{andrieu2004particle}
C.~Andrieu, A.~Doucet, S.~Singh, and V.~Tadic, ``Particle methods for change
  detection, system identification, and control,'' \emph{Proc. IEEE}, vol.~92,
  no.~3, pp. 423--438, 2004.

\bibitem{rezende2015variational}
D.~Rezende and S.~Mohamed, ``Variational inference with normalizing flows,'' in
  \emph{Proc. Int. Conf. Mach. Learn. (ICML)}, Lille, France, July 2015.

\bibitem{burda2016importance}
Y.~Burda, R.~B. Grosse, and R.~Salakhutdinov, ``Importance weighted
  autoencoders,'' in \emph{Proc. Int. Conf. Learn. Rep. (ICLR)}, San Juan,
  Puerto Rico, May. 2016.

\bibitem{jordan1999introduction}
M.~I. Jordan, Z.~Ghahramani, T.~S. Jaakkola, and L.~K. Saul, ``An introduction
  to variational methods for graphical models,'' \emph{Mach. Learn.}, vol.~37,
  no.~2, pp. 183--233, 1999.

\bibitem{beal2003variational}
M.~J. Beal, \emph{Variational algorithms for approximate {B}ayesian
  inference}.\hskip 1em plus 0.5em minus 0.4em\relax University of London,
  University College London, United Kingdom, 2003.

\bibitem{rezende2014stochastic}
D.~J. Rezende, S.~Mohamed, and D.~Wierstra, ``Stochastic backpropagation and
  approximate inference in deep generative models,'' in \emph{Proc. Int. Conf.
  Mach. Learn. (ICML)}, Beijing, China, June 2014.

\end{thebibliography}
	
\end{document}